%% file: sample-sigconf-authordraft.tex
\documentclass[sigconf]{acmart}

\makeatletter
\def\@ACM@checkaffil{
    \if@ACM@instpresent\else
    \ClassWarningNoLine{\@classname}{No institution present for an affiliation}%
    \fi
    \if@ACM@citypresent\else
    \ClassWarningNoLine{\@classname}{No city present for an affiliation}%
    \fi
    \if@ACM@countrypresent\else
        \ClassWarningNoLine{\@classname}{No country present for an affiliation}%
    \fi
}
\makeatother

\AtBeginDocument{%
  }

\setcopyright{acmlicensed}
\copyrightyear{2018}
\acmYear{2018}
\acmDOI{XXXXXXX.XXXXXXX}

\acmConference[Conference acronym 'XX]{Make sure to enter the correct
  conference title from your rights confirmation emai}{June 03--05,
  2018}{Woodstock, NY}
\acmISBN{978-1-4503-XXXX-X/18/06}

\usepackage{makecell}
\usepackage{multirow}
\usepackage{multicol}
\usepackage[normalem]{ulem}
\useunder{\uline}{\ul}{}
\usepackage{hyperref}
\usepackage{threeparttable}
\usepackage{algorithm}
\usepackage{algpseudocode}
\usepackage{amsmath,amsfonts}
\usepackage{float}
\usepackage{enumitem}
\usepackage{graphicx}
\usepackage{subcaption}
\usepackage{afterpage}
\usepackage{stfloats}

\setcopyright{acmlicensed}

\copyrightyear{2024}
\acmYear{2024}
\acmDOI{XXXXXXX.XXXXXXX}

\acmConference[SIGSPATIAL '24]{32nd ACM SIGSPATIAL International Conference on Advances in Geographic Information Systems}{October 29 - November 1, 2024}{Atlanta, GA, USA}
\acmISBN{978-1-4503-XXXX-X/18/06}

\begin{document}

\title{Enhancing Spatio-temporal Quantile Forecasting with Curriculum Learning: Lessons Learned}


\author{Du Yin}
\email{du.yin@unsw.edu.au}
\affiliation{
  \institution{University of New South Wales\\Sydney, NSW, Australia}
}

\author{Jinliang Deng}
\email{dengjinliang@ust.hk}
\affiliation{
  \institution{HKGAI, Hong Kong SAR\\ Southern University of Science and Technology, China}
}

\author{Shuang Ao}
\email{shuang.ao@unsw.edu.au}
\affiliation{
  \institution{University of New South Wales\\Sydney, NSW, Australia}
}

\author{Zechen Li}
\email{zechen.li@unsw.edu.au}
\affiliation{
  \institution{University of New South Wales\\Sydney, NSW, Australia}
}

\author{Hao Xue}
\email{hao.xue1@unsw.edu.au}
\affiliation{
  \institution{University of New South Wales\\Sydney, NSW, Australia}
}

\author{Arian Prabowo}
\email{arian.prabowo@unsw.edu.au}
\affiliation{
  \institution{University of New South Wales\\Sydney, NSW, Australia}
}

\author{Renhe Jiang}
\email{jiangrh@csis.u-tokyo.ac.jp}
\affiliation{%
  \institution{Center for Spatial Information Science, The University of Tokyo}
}

\author{Xuan Song}
\email{songxuan@jlu.edu.cn}
\affiliation{
  \institution{School of Artificial Intelligence of Jilin University}
}

\author{Flora Salim}
\email{flora.salim@unsw.edu.au}
\affiliation{
  \institution{University of New South Wales\\Sydney, NSW, Australia}
}

\renewcommand{\shortauthors}{Trovato et al.}

\begin{abstract}
Training models on spatio-temporal (ST) data poses an open problem due to the complicated and diverse nature of the data itself, and it is challenging to ensure the model's performance directly trained on the original ST data. While limiting the variety of training data can make training easier, it can also lead to a lack of knowledge and information for the model, resulting in a decrease in performance. 
To address this challenge, we presented an innovative paradigm that incorporates three separate forms of curriculum learning specifically targeting from spatial, temporal, and quantile perspectives. Furthermore, our framework incorporates a stacking fusion module to combine diverse information from three types of curriculum learning, resulting in a strong and thorough learning process. We demonstrated the effectiveness of this framework with extensive empirical evaluations, highlighting its better performance in addressing complex ST challenges. We provided thorough ablation studies to investigate the effectiveness of our curriculum and to explain how it contributes to the improvement of learning efficiency on ST data.
\end{abstract}

\ccsdesc[500]{Do Not Use This Code~Generate the Correct Terms for Your Paper}
\ccsdesc[300]{Do Not Use This Code~Generate the Correct Terms for Your Paper}
\ccsdesc{Do Not Use This Code~Generate the Correct Terms for Your Paper}
\ccsdesc[100]{Do Not Use This Code~Generate the Correct Terms for Your Paper}

\keywords{Deep learning, Spatio-temporal, Curriculum learning, Quantile forecasting}

\received{20 February 2007}
\received[revised]{12 March 2009}
\received[accepted]{5 June 2009}


\maketitle

\section{Introduction}

Traffic forecasting~\cite{jiang2021dl} endeavors to anticipate future traffic patterns in road networks by analyzing historical data. However, addressing the task of training a deep learning model in complex and diverse ST data presents a prominent challenge.  
Recent work has demonstrated the success of the model's capacity to capture the inherent spatio-temporal dependencies in traffic prediction \cite{wu2019graph, wu2020connecting,diao2019dynamic,shang2021discrete,ye2021coupled,jiang2023spatio, shao2023exploring}. 
The underlying premise of spatio-temporal graph modeling is that the future information of a node relies on its historical information as well as its neighboring nodes. Quantile forecasting models such as MQRNN~\cite{wen2017multi}, DeepAR~\cite{salinas2020deepar}, and TFT~\cite{lim2021temporal} aim to provide the distribution of observations for each prediction horizon, characterized by the upper and lower limits of probable observations. Despite the improved performance, most efforts have concentrated on optimizing the network architecture, with less attention paid to the training procedure, which is also of paramount importance to a competent and robust model.

\begin{figure}[h!]
  \centering
  \includegraphics[width=\linewidth]{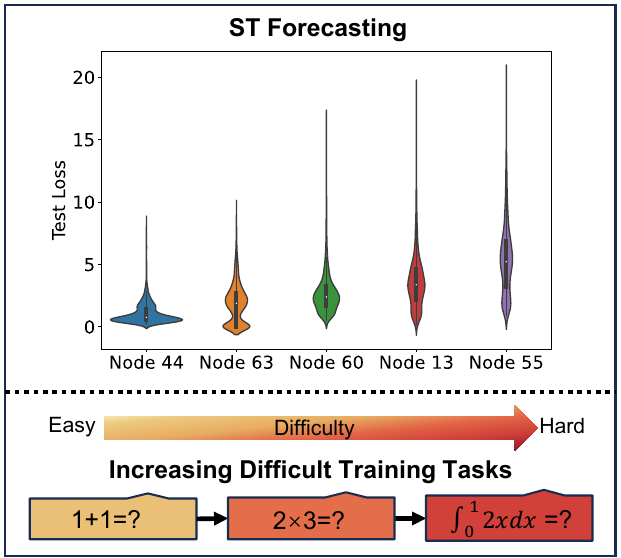}
  \caption{Spatio-temporal problems often show a clear separation between easy and difficult instances, closely mirroring the curriculum learning approach that progresses from simple to complex tasks. This similarity suggests that curriculum learning's gradual difficulty increase is well-suited for addressing the varied complexities of spatio-temporal forecasting.}
  \label{fig1}
\end{figure}

The concept of curriculum learning (CL) emerges as a promising option to achieve high-quality and efficient training, where the constructed model is trained on instances of increasing difficulty, mimicking the learning process of human beings~\cite{bengio2009curriculum}. More specifically, CL commences with easy instances, and then gradually escalates the complexity of the considered instances until encompassing the entire training data. This progress is managed by a scheduler, which determines the order and pace of introducing training instances based on their difficulty level. An effective scheduler ensures a balanced learning process, preventing the model from being overwhelmed by overly difficult instances at the beginning and avoid under-challenged by overly easy instances during the training.
Since its inception, CL has exhibited its advantages on many challenging tasks~\cite{wang2021survey}. 

However, conventional instance-level CL scheduler does not fit the unconventional sampling method adopted in spatial-temporal forecasting tasks, resulting in considerable unnecessary computations. To combat with this issue, we propose a novel group-level scheduler, evaluating the challenge for handling a group of instances collectively driven by a common factor. Motivated by the fact that hard instances tend to occur simultaneously across different variables or repeatedly over the same variable across different times as a result of spatio-temporal correlations, as illustrated in Figure~\ref{fig1}, we consider the factors of spatial identity and temporal identity, leading to a spatial-view scheduler and a temporal-view scheduler, respectively. Encoding these two identities has been proven paramount for promoting the forecast accuracy~\cite{shao2022spatial, deng2021st}. Distinct from previous methods, our study explores how these two identities can benefit the training quality and efficiency, eliminating unnecessary computations. In addition, to deal with quantile forecasting, we invent a quantile-view  scheduler. The experimental results substantiate the efficacy of our curriculum in accommodating diverse methodologies and improving their performance. Furthermore, we offer comprehensive experiments and assessments of the insights we gained regarding the curriculum's design. Our contribution can be summarized as follows:

\begin{itemize}[leftmargin=*]
    \item We innovatively formulate the spatial-temporal quantile forecasting problem within the framework of curriculum learning.
    \item We propose Spatio-Temporal-Quantile Curriculum Learning (STQCL), a framework with three specialized curriculum learning schedulers for spatial, temporal, and quantile data. STQCL utilizes three schedulers to simplify the training process, and integrates all three data types, demonstrating STQCL's \textbf{adaptability to any ST models}.
    
    
    \item We compare STQCL with different curriculum designed for spatio-temporal-quantile data, and summarize the insights and lessons learned from curriculum structure design. Evaluating STQCL in six popular datasets, we demonstrate its performance improvement of both point and quantile forecasting across all basic ST models.
    
\end{itemize}


\section{Related Work}
\subsection{Spatio-temporal Forecasting Overview}
Spatio-temporal forecasting is a longstanding challenge over the last few decades. Conventionally, statistical models like AR~\cite{hamilton1994autoregressive}, VAR~\cite{stock2001vector}, and ARIMA~\cite{pan2012utilizing} were widely used in this domain. The recent advent of deep learning models has significantly transformed the landscape,  dominating the traditional statistical methods.

The high compatibility of traffic graph data with graph models inspired numerous excellent studies. For example, T-GCN~\cite{zhao2019t} merged graph convolution with GRU~\citep{chung2014empirical}, altering GRU~\cite{chung2014empirical} cell computations to graph convolutions. STGCN~\cite{yu2018spatio} introduced a Chebyshev-based graph convolution and a spatio-temporal convolutional stack. DCRNN~\cite{li2018diffusion} used a diffusion convolution based on random walks and an encoder-decoder framework. GWN~\cite{wu2019graph} proposed a adaptive graph learning method coupled with a WaveNet~\cite{oord2016wavenet} architecture. Subsequent innovations have been built on these foundational works. For instance, ASTGCN~\cite{guo2019attention} enhanced STGCN by incorporating attention mechanisms and adding more temporal features, such as hour, day, and week statistical characteristics. T-MGCN~\cite{lv2020temporal} improved upon the TGCN framework by integrating multi-graph concepts. MegaCRN~\cite{jiang2023spatio} extended DCRNN's~\cite{li2018diffusion} backbone by incorporating graph learning and meta-learning. In parallel with the above advancements, STNorm~\cite{deng2021st}, MTGNN~\cite{wu2020connecting}, and DMSTGCN~\cite{han2021dynamic} managed to enhance GWN~\cite{wu2019graph}, showcasing the continuous evolution and application of these advanced techniques in spatio-temporal forecasting.

Transformer, proven as a powerful backbone across various fields in recent years, has also similarly catalyzed innovative works in the spatio-temporal domain. For instance, STGNN~\cite{wang2020traffic} and PDFormer~\cite{pdformer} integrated the spatial characteristics of spatio-temporal problems into the Transformer framework, ingeniously embedding or enhancing spatial feature extraction within a robust temporal feature extraction backbone.

Spatio-temporal quantile forecasting gains increasing attention over the last few years. Early works like MQRNN~\cite{wen2017multi}, DeepAR~\cite{salinas2020deepar}, and TFT~\cite{lim2021temporal} have explored quantile forecasting within general time series prediction, combining deep learning with probabilistic distribution techniques to tackle the uncertainty in time series. However, these models were not specifically designed for spatio-temporal issues, as their primary focus remained on temporal features. \citet{wu2021quantifying} conducted a systematic study of uncertainty quantification methods for spatiotemporal forecasting, including quantile regression.

Our work develops a novel CL paradigm to support both point forecasting and uncertainty forecasting (in this work, uncertainty is represented through quantile forecasting) in its outputs, bridging a gap in the current research landscape.


\subsection{Curriculum Learning Overview}
Curriculum learning is well-adopted beyond specific machine learning tasks, e.g., natural language processing (NLP)~\cite{Tay2019SimpleAE}, computer vision (CV)~\cite{Guo2018CurriculumNetWS}, reinforcement learning (RL)~\cite{copilot_NEURIPS2021_56577889, Portelas2019TeacherAF}, as well as graph learning~\cite{Gong2019MultiModalCL} and neural architecture search (NAS)~\cite{Guo2020BreakingTC}. In the context of spatio-temporal forecasting, a simple practice of CL has been integrated into the training process by previous methods~\cite{li2018diffusion, wu2020connecting, guo2021learning, shao2022pre, shao2022decoupled}, which gradually increase the prediction length by every CL step size. However, the contribution of CL to performance improvement in these models is found to either negligible or limited, suggesting the need for further exploration in the adaptation of CL to spatio-temporal forecasting.

\section{preliminary}

\subsection{Definition of Spatio-temporal Quantile Forecasting}
Following the previous work~\cite{jiang2021dl}, a spatio-temporal problem is defined as follows: 
\begin{equation}
[X_{t-(T_\text{in}-1)},...,X_{t-1},X_{t}] \rightarrow [X_{t+1},X_{t+2},...,X_{t+T_\text{out}}].
\end{equation}
where $X_i$ $\in$ $\mathbb{R}^{N\times C}$ refers to the time-series tensor of the \(i^{th}\)  step, $N$ represents the number of the ST data (e.g., sensors or roads) and $C$ is the feature dimension of ST data. $T_\text{in}$ and $T_\text{out}$ denote the lengths of input and output sequences, respectively.

Traditionally, spatio-temporal issues have predominantly been approached through point predictions, with \(C\) generally set to 1. In our endeavor, we facilitate the simultaneous output of point predictions and quantile predictions by setting \(C\) to 3. This straightforward method allows for the differentiation between the upper boundary, median (equivalent to the output of point predictions), and lower boundary:
\begin{equation}
\label{qloss}
l_k(X, \hat{X})= \begin{cases}\alpha \cdot(X-\hat{X}) &  \hat{X} \leq X \\ (1-\alpha) \cdot(\hat{X}-X) &  \hat{X}>X\end{cases}.  
\end{equation}
where $X$ and $\hat{X}$ represent the objective values and the prediction values, respectively; $\alpha$ denotes the quantile to be forecasted, with the fitted model tending to overestimate results in approximately \((\alpha \times 100)\%\) of cases and underestimate in \((100 - \alpha \times 100)\%\) of cases.

\subsection{Curriculum Learning}

Curriculum Learning (CL), inspired by natural learning sequences as discussed by~\cite{bengio2009curriculum}, enhances deep learning model performance by progressively introducing complexity through a structured sequence of gradually increasing difficulty, demonstrating a more efficient learning process. Spatio-temporal problems vary in difficulty, which fits well with curriculum learning's step-by-step complexity increase. This inspired us to use curriculum learning in spatio-temporal forecasting to gradually address its complexity.


Algorithm~\ref{cl_algorithm} shows the basic steps for training a model $M$ using curriculum learning on dataset $D$. It requires a curriculum strategy to decide when, where, and in what order to apply the data for model training, such as at the data level, model level, or performance measurement level evaluated by $g$. The scheduler 
determines the pace solely based on the current training iteration/epoch, using the grades calculated by $C$ and guiding the subsequent curriculum training strategy.

\begin{algorithm}
\caption{Simple Curriculum Learning Algorithm}
\label{cl_algorithm}
\begin{algorithmic}[1]
\Statex  $M \gets$ a deep learning model;
\Statex  $D \gets$ a training data set;
\Statex  $C \gets$ curriculum calculation and scheduler;

\While {training:}
    \State get $g \gets$ $C(M,D)$
    \State update $D' \gets$ select$(D)$
    \State update $M' \gets$ train$(M, D', g)$
\EndWhile
\end{algorithmic}
\end{algorithm}

\section{Spatio-Temporal Quantile Curriculum Learning}
\subsection{Overview}

Our spatio-temporal quantile curriculum learning framework primarily builds upon state-of-the-art (SOTA) models for existing spatio-temporal problems. In this section, we will detail our method by illustrating how and when to implement curriculum learning. We adapt self-paced learning (SPL)~\cite{kumar2010self} from the curriculum learning branch to serve as a component within spatio-temporal models. Targeting the difficulty level assessment as performed by the $G$ module in Algorithm~\ref{cl_algorithm}, we utilize evaluation metrics specific to spatio-temporal challenges to further guide the adjustment of the curriculum's difficulty level.

Different from the tasks in image classification, text recognition, and other tasks in the fields of computer vision (CV) and natural language processing (NLP), spatio-temporal forecasting problems often lack reliable prior knowledge. Therefore, traditional curriculum learning methods that rely on predefined difficulty measurer based on prior knowledge before training~\cite{wang2021survey} may not be suitable for spatio-temporal prediction tasks. SPL, which assesses difficulty using data or the model itself, fits this task well. 

\begin{figure*}[h!]
  \centering
  \includegraphics[width=\linewidth]{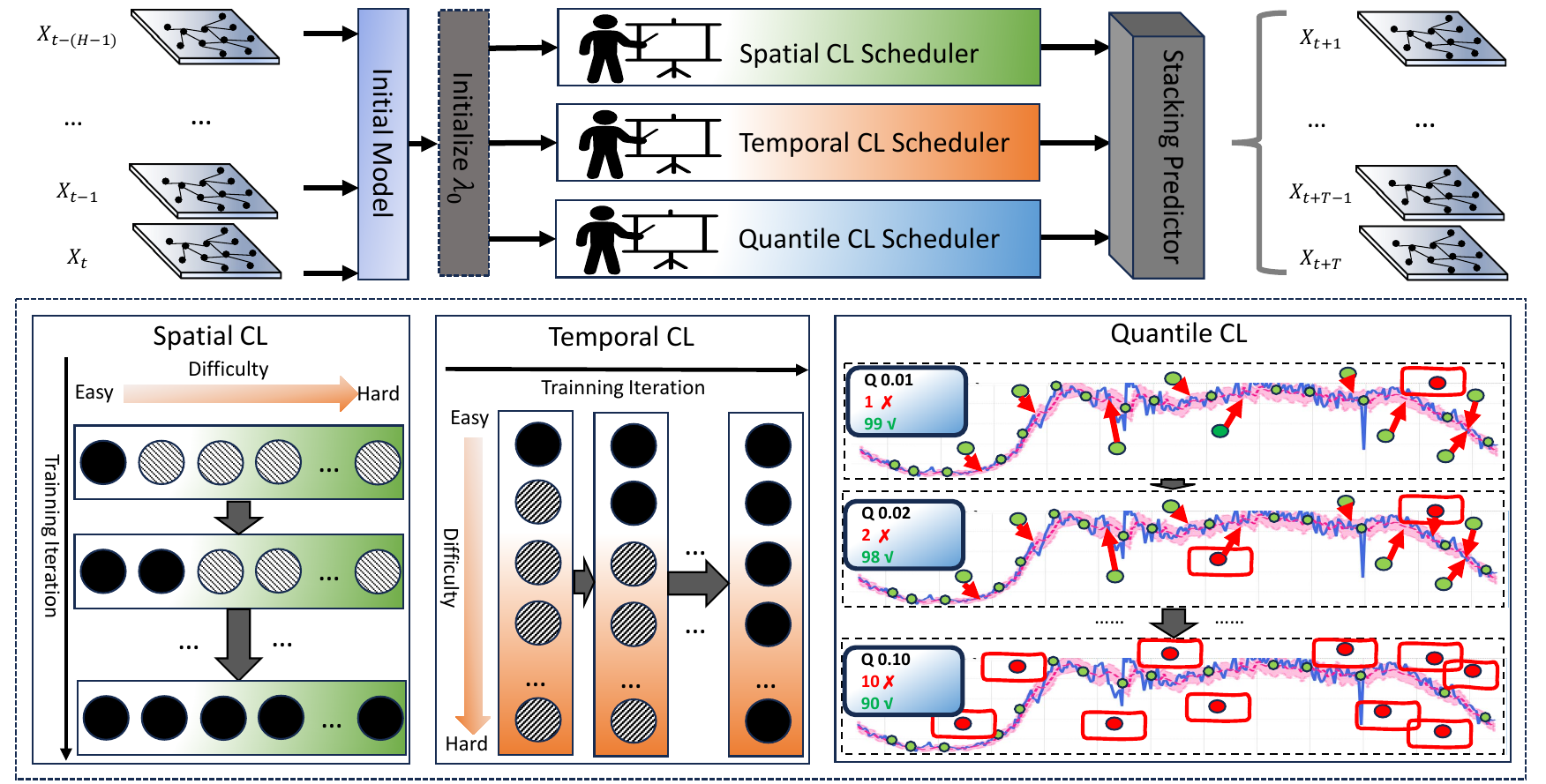}
  \caption{The overview of our spatio-temporal-quantile curriculum learning framework. First, ST data is trained to get an initial model for initializing the curriculum learning. Subsequently, the three types of curriculum learning schedulers proposed in this work are applied. The knowledge from the three curriculum learning experts is then fused using a simple linear layer. The details of each type of curriculum learning are illustrated in the latter part of the figure.}
  \label{fig2}
\end{figure*}

As shown in Figure~\ref{fig2}, SPL based on the loss function initially creates a difficulty ranking with an initial model. The model serves as its own teacher, using difficulty scores derived from this initial assessment to guide subsequent curriculum learning and training intervention. Building upon this foundation, our study introduces three curriculum learning strategies (spatial, temporal, and quantile) to extract features from different perspectives. Finally, a stacking fusion module integrates the advantages of all three strategies, enhancing the predictive model's performance.

\subsection{Problem Formulation within Curriculum Learning}
Spatio-temporal quantile forecasting is originally formulated as follows:
\begin{equation}
\min _{\boldsymbol{w}}  \sum_{i=1}^N \sum_{j=1}^T \sum_{k=1}^Q l_{k} \left(f\left(X_{j-T_\text{in}+1:j, i}; \boldsymbol{w}\right), X_{j+1:j+T_\text{out}, i}\right) \label{eq:loss}.
\end{equation}
where $\boldsymbol{w}$ denotes the model parameters and $l_k$ denotes the quantile loss against the k$^\text{th}$ quantile. For brevity, we use $l_{ijk}$ to represent $l_{k} \left(f\left(X_{j-T_\text{in}+1:j, i}; \boldsymbol{w}\right), X_{j+1:j+T_\text{out}, i}\right)$ below.

We then extend the formulation of SPL to spatial-temporal quantile forecasting problems. Following the framework of SPL~\cite{kumar2010self}, problem \ref{eq:loss} is modified by introducing a tensor of binary variables $\boldsymbol{v}$ $\in$ $\{0, 1\}^{N\times T\times Q}$, each of which indicates whether the corresponding instance is easy or not:
\begin{equation}
\min _{\boldsymbol{w} ; \boldsymbol{v} \in[0,1]^{STQ}} \sum_{i=1}^N \sum_{j=1}^T \sum_{k=1}^Q \boldsymbol{v}_{ijk} \boldsymbol{l}_{ijk} -\lambda \sum_{i=1}^N \sum_{j=1}^T \sum_{k=1}^Q \boldsymbol{v}_{ijk} \label{loss_new}.
\end{equation}
where $\lambda$ stands for the age parameter that decides the selected instances at the current training iteration, controlling the learning pace of the training progress. Specifically, by progressively increasing the value of $\lambda$, a growing number of difficult samples are included in the training process until all the samples are considered. \citet{kumar2010self} demonstrated that this problem can be efficiently solved by alternative convex search (ACS)~\cite{bazaraa2013nonlinear}, which alternately optimizes $\boldsymbol{w}$ and $\boldsymbol{v}$ while keeping the other set of variables fixed. In particular, given parameters $\boldsymbol{w}$, we can obtain the optimum of $\boldsymbol{v}$ via:
\begin{equation}
\label{vv}
\boldsymbol{v}_{ijk}=\left\{\begin{array}{lr}1, & l_{ijk}<\lambda \\ 0, & \text { otherwise }\end{array}\right..
\end{equation}
If $\lambda$ grows over the maximum of $l_{ijk}$, all the instances are taken into consideration. Then, for every given set of $\boldsymbol{v}$, problem~\ref{loss_new} is reduced to the following form to optimize $\boldsymbol{w}$:
\begin{equation}
\label{loss_new_new}
\min _{\boldsymbol{w}}  \sum_{i=1}^N \sum_{j=1}^T \sum_{k=1}^Q \boldsymbol{v}_{ijk} \boldsymbol{l}_{ijk}.
\end{equation}

It is important to note that initiating curriculum learning requires a substantial number of courses. Therefore, our first step is to initialize $\boldsymbol{w}$ and $\boldsymbol{v}$, and set $\lambda$ to $\lambda_0$ (where $\lambda_0$ is determined by the initial model). Thus,  the algorithm is implemented as Algorithm~\ref{cl2_algorithm}.

\begin{algorithm}
\caption{Self-paced Spatio-temporal Quantile Curriculum Learning Training Algorithm}
\label{cl2_algorithm}
\begin{algorithmic}[1]
\Statex \textbf{Input:} $M$: the deep learning model; $D$: the training datasets; $n$: training epoch/iteration; $\mu$: CL stepsize
\Statex \textbf{Output:} $\boldsymbol{w}$: The optimal parameters of $M$
\State Initialize $\boldsymbol{w}$, $\boldsymbol{v}$, $\lambda$=$\lambda_0$, $i$=1
\For{$i = 1$ to $n$}
\If{$i \% \mu = 0$}
\State update $\boldsymbol{v}$ by Eq.~\ref{vv};
\State update $\boldsymbol{\lambda}$ to a larger value; 
\EndIf
\State update $\boldsymbol{w}$ by Eq.~\ref{loss_new_new}; 
\EndFor
\end{algorithmic}
\end{algorithm}

\subsection{Spatio-temporal-quantile Curriculum Learning Scheduler}
\label{sec43}

The conventional instance-level scheduler deals with each instance independently, failing to offer optimal efficiency for spatial-temporal forecasting problems. Motivated by the spatio-temporal correlations present in data, we design three types of group-level curriculum learning schedulers to update $\boldsymbol{v}$: spatial curriculum learning scheduler, one temporal, and one quantile.

We first give an explanation of why instance-level scheduler is suboptimal in terms of computational efficiency, which is in closely related to the instance selection scheme utilized to train the model. For the majority of machine learning problems, random sampling is adopted as the default choice with the assumption that data points are independent and identically distributed. When encountering hard instances identified by the instance-level scheduler, prior methods can simply discard them and resample new instances until a batch is filled. Contrary to the i.i.d. assumption, spatial-temporal forecasting problems involve data showcasing spatial and temporal correlations. Therefore, stratified sampling is applied in place of random sampling. Stratified sampling considers groups of data points collected at the same time, i.e., $X_t$, simultaneously. Akin to random sampling, it also runs in a batch manner, iteratively yielding a batch of instance groups, i.e. $\{X_{t_1}, \cdots, X_{t_b}\}$, organized in a matrix of instances. To maximize the performance of parallel computation, the data points associated with the same spatial identity are required to be aligned across the groups within the same batch. As a consequence, instance-level scheduler could leave "gaps" in arbitrary positions of the instance matrix for each iteration of training. Moreover, these gaps cannot be shrunk or replaced, incurring a substantial amount of unnecessary computation spent on them, especially at the early stage of training, leading to undesirable efficiency loss. 

The spatial-view scheduler assigns each variable an individual difficulty score, which measures the overall degree of difficulty for that variable based on its entire history. Then, the indicators derived by the spatial-view scheduler are calculated in the following way:
\begin{equation}
\boldsymbol{v}_{ijk}=\left\{\begin{array}{lr}1, & \frac{1}{TQ}\sum_j\sum_k l_{ijk}<\lambda_s \\ 0, & \text{otherwise }\end{array}\right..
\end{equation}
The ranking of difficulty scores among the variables determines the timing they are incorporated into the training process. Over the course of the training process, for those variables deemed excluded, the scheduler can shrink the corresponding columns without leaving any vacuums.

The similar idea applies to the temporal-view scheduler, computing the average score across the variables to determine the indicators:
\begin{equation}
\boldsymbol{v}_{ijk}=\left\{\begin{array}{lr}1, & \frac{1}{NQ}\sum_i\sum_k l_{ijk}<\lambda_t \\ 0, & \text { otherwise }\end{array}\right..
\end{equation}
In this way, an entire row of instances is either valid or invalid, allowing the resampling additional rows of data to fill in the invalid ones. 

The scheduler for the quantile part differs slightly. We no longer use the hard regularizers from the spatial scheduler. Instead, we adjust the difficulty and simplicity by iteratively changing the boundaries in the quantile loss function:

\begin{equation}
\boldsymbol{v}_{ijk}=\left\{\begin{array}{lr}\alpha_{k}, & \frac{1}{NT}\sum_i\sum_j l_{ijk}<\lambda_q \\ 0, & \text { otherwise }\end{array}\right..
\end{equation}

where $\alpha$ adjusts the quantile boundaries of the quantile loss function. Refer to Figure~\ref{fig2} for more details.


\subsection{Stacking Fusion Predictor}

To harness the capabilities of the three types of curriculum learning schedulers, we developed a stacking fusion predictor module. This module merges the outputs from the three schedulers and processes them through a simple linear layer to generate the final quantile predictions.
\begin{equation}
\mathcal{Z}=\operatorname{Linear}\left(\operatorname{Concat}\left(\mathcal{O}_{s}, \mathcal{O}_{t}, \mathcal{O}_{q}\right) \in \mathbb{R}^{N \times T \times Q*3}\right).
\end{equation}

\section{Experiments}
\subsection{Settings}

\subsubsection{Datasets}
In order to validate the proposed method, we conducted experiments on six of the most popular datasets: PEMS03, PEMS04, PEMS07, and PEMS08\footnote{\url{https://pems.dot.ca.gov/}} \{released by STSGCN~\cite{song2020spatial}\} and METR-LA, PEMSBAY \{released by DCRNN~\cite{li2018diffusion}\}. The information about these datasets is summarized in Table~\ref{data}. We selected PEMS04 and METR-LA to showcase the comparison of baselines in this section. Additional datasets, including PEMS03, PEMS07, PEMS08, and PEMSBAY, are presented in the Appendix~\ref{appendix}.

\input{table/data.tex}

\input{table/main_experiment.tex}

\subsubsection{Baselines}


We embedded the STQCL model into some of the most popular state-of-the-art (SOTA) models, conducted extensive experiments on two open datasets, and achieved enhancements in both point and quantile prediction metrics. As discussed in related work, we've selected models representing diverse architectures to showcase our method's robustness: STGCN~\cite{yu2018spatio} for its sandwich structure, DCRNN~\cite{li2018diffusion} for its Encoder-Decoder framework, GWN~\cite{wu2019graph} for its WaveNet and learnable graph model, SCINet~\cite{liu2022scinet} for its stacked MLP architecture, STAEFormer~\cite{liu2023spatio} leveraging Transformer as its backbone, and STNorm~\cite{deng2021st} for its normalized flow feature enhancement. Choosing methods across different frameworks and technologies as basic models strongly validates the effectiveness of our approach.



\subsubsection{Evaluation Metrics}

Root Mean Square Error (RMSE), Mean Absolute Error (MAE), and Mean Absolute Percentage Error (MAPE) are adopted as evaluation metrics for point forecasting, being the most prevalent indicators in traffic point forecasting. To quantify uncertainty in our forecasting, we follow the work such as MQRNN~\cite{wen2017multi} and utilize Quantile Loss as our evaluation metric. Furthermore, we specify the quantiles for upper, median, and lower prediction as \{0.1, 0.5, and 0.9\}, respectively.


\subsubsection{Implementation Details}


Our experiment was conducted using PyTorch, with the Adam optimizer with Quantile Loss as the loss function. The learning rate was set to $1e^{-3}$. Additionally, an early stopping strategy was implemented with a patience of 10 epochs to prevent overfitting. To mitigate the impact of random seeds, the experiment was replicated five times, with results averaged across these iterations. In all experiments, the input sequence consisted of 12 steps, with a prediction span of 12 steps as well. Evaluation was specifically focused on the $3^{rd}$, $6^{th}$, and $12^{th}$ steps of the output.
The data was divided into training, validation, and testing datasets in different proportions for two datasets: 70\% training, 10\% validation, and 20\% testing for the METR-LA dataset, and 60\% training, 20\% validation, and 20\% testing for the PEMS04 dataset. The code of the implementation is available at the anonymous repositories:\href{https://github.com/cruiseresearchgroup/STQCL}{https://github.com/cruiseresearchgroup/STQCL} .

\subsection{Results}

\subsubsection{Overall Comparison}
Table~\ref{main_experiment} presents the results for all baseline models and their enhancements using our proposed STQCL framework. The superior results are highlighted with bold and underlined formatting. From these results, we can deduce the following: (1) Our STQCL approach is applicable to all state-of-the-art (SOTA) models, enhancing their capabilities to varying extents. This demonstrates the versatility and effectiveness of STQCL in leveraging the potential of existing models. (2) Crucially, providing upper and lower boundaries to represent uncertainty is of significant importance. So transforming traditional point prediction traffic models into a combination of point prediction and quantile prediction is a viable contribution.  As demonstrated in Table~\ref{point}, our dual-output model, when compared to traditional point forecast models, reveals that our modifications not only enable additional quantile forecasting but also enhance point prediction performance. (3) More complex models, such as STAEFormer~\cite{liu2023spatio} and GWN~\cite{wu2019graph}, exhibit substantial improvements, indicating that the efficacy of our STQCL strategy is correlated with the complexity of the models. This suggests that STQCL not only complements but also maximizes the potential of sophisticated SOTA models. Additionally, we selected two quantile prediction baselines, DeepAR and MQRNN, for comparison purely based on the Q-loss metric. As Table~\ref{quantile_com} illustrates, our modified ST model for quantile prediction outperforms traditional quantile forecasting models, and our STQCL further enhances its quantile prediction performance.

\input{table/point}
\input{table/quantile}

\subsubsection{Case Study}

\begin{figure}[h!]
  \centering
  \begin{subfigure}[b]{0.495\linewidth}
    \centering
    \includegraphics[width=\linewidth]{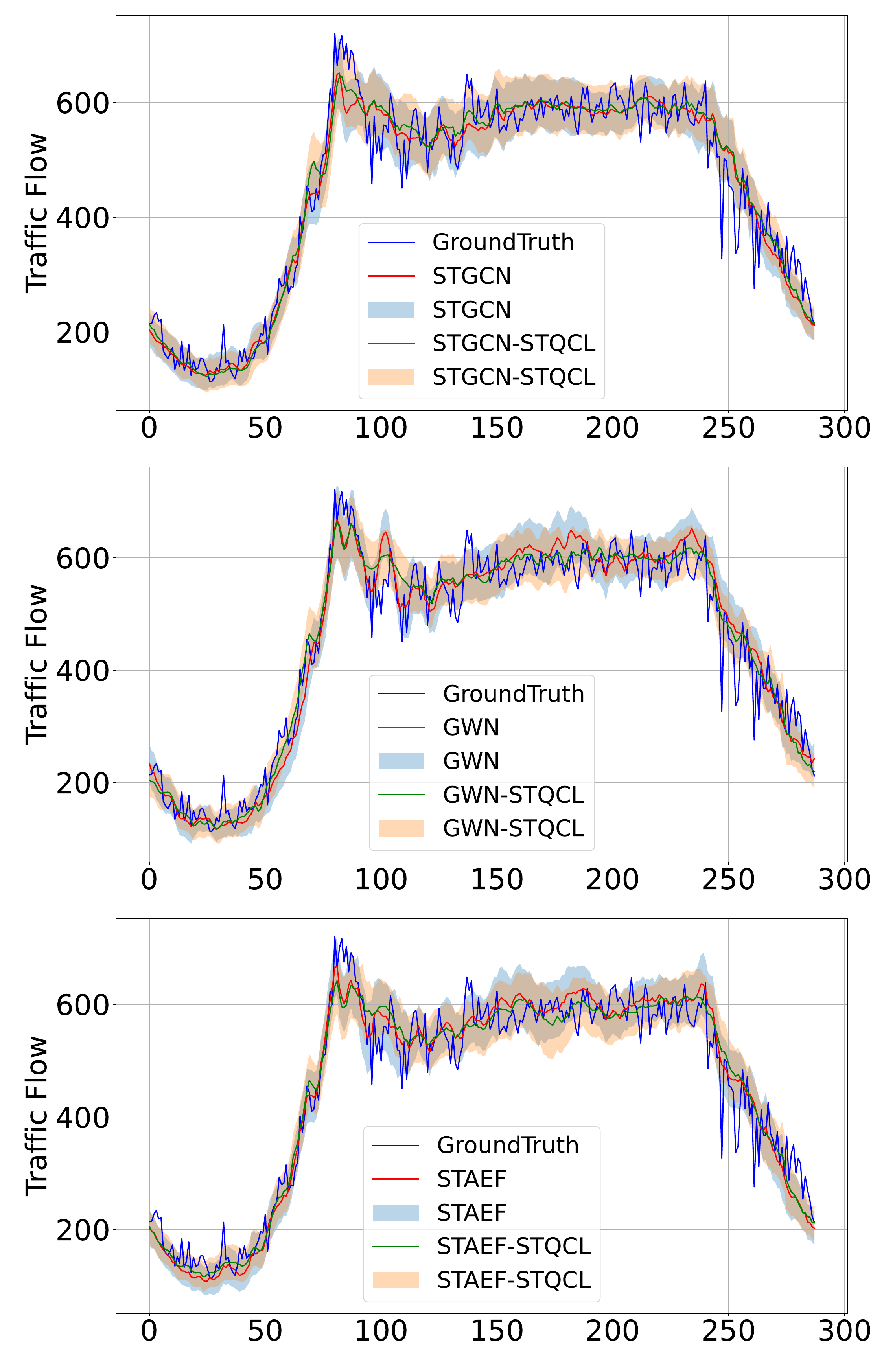}
    \caption{Case in sensor 58}
    \label{case1:sub1}
  \end{subfigure}
  \hfill
  \begin{subfigure}[b]{0.495\linewidth}
    \centering
    \includegraphics[width=\linewidth]{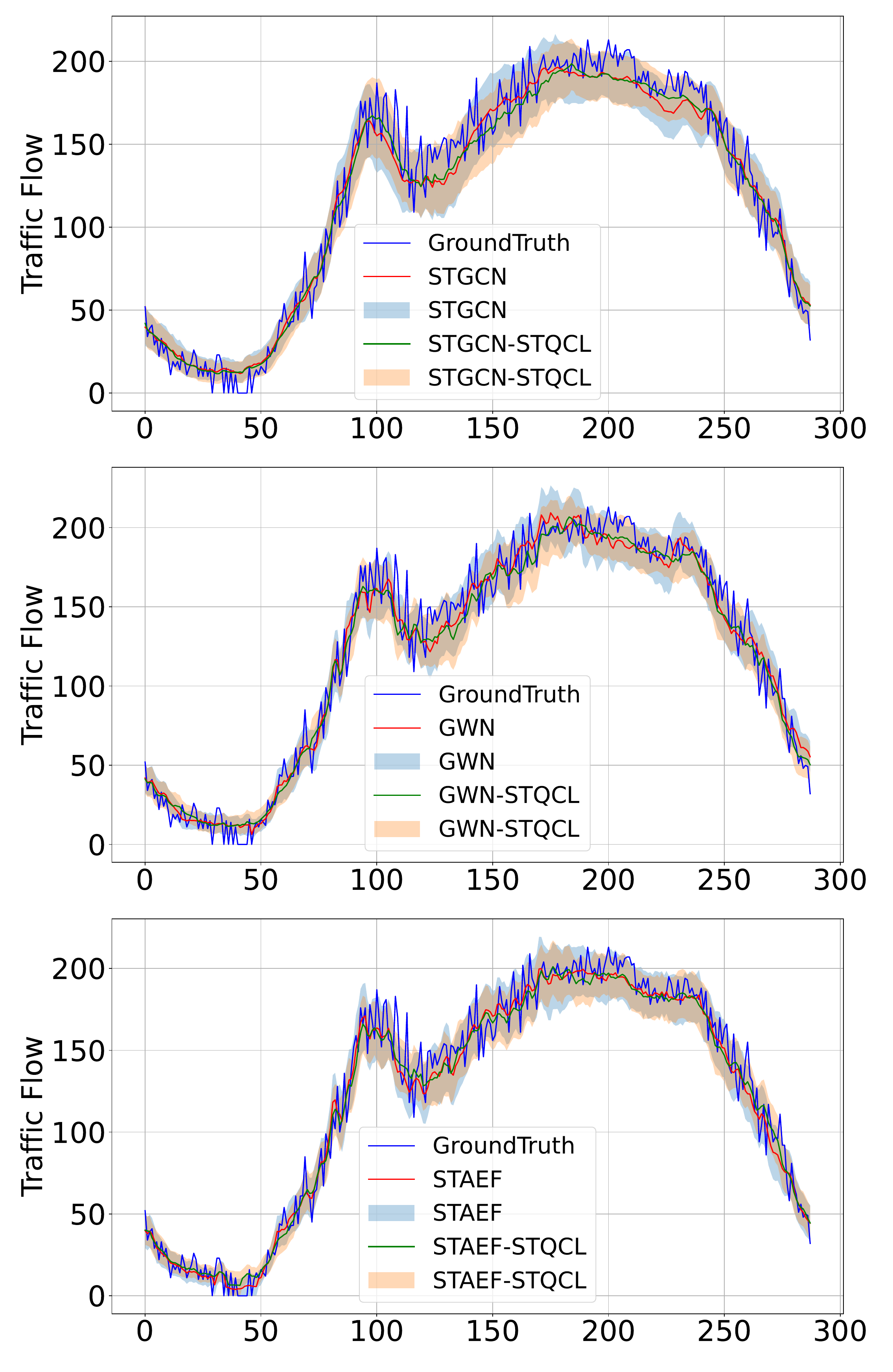}
    \caption{Case in sensor 112}
    \label{case1:sub2}
  \end{subfigure}
  \caption{Case study on PEMS04.}
  \label{case1}
\end{figure}

We randomly selected two sensors from the PEMS04 dataset to analyze the most challenging forecasting scenario with a horizon of 12. As illustrated in Figure~\ref{case1}: the integration of our STQCL framework has markedly enhanced the performance of basic model. Whether it's the precise point predictions (illustrated by solid lines) or within the range of uncertainty in predictions (i.e., quantile predictions, indicated by shaded areas), the enhanced model shows notable improvements over the baseline models. These advancements not only underscore the potent capabilities of STQCL framework but also lay the groundwork for further research and application.

\subsubsection{Ablation Study}

An ablation study was also conducted to compare the three curriculum learning modules proposed in this work. These include: (1) Spatial-Curriculum Learning (SCL), which progressively trains on samples from easy to difficult from a spatial perspective; (2) Temporal-Curriculum Learning (TCL), focusing on the temporal dimension; and (3) Quantile-Curriculum Learning (QCL), representing the difficulty implied by quantile boundaries. We analyzed the results for a horizon of 12 on STGCN and its variants enhanced with different Curriculum Learning (CL) modules. The results, as reported in Table~\ref{ablation}, show that each module contributes to marginal improvements. We conclude that integrating different curriculum learning modules further enhances performance.
\input{table/ablation}

\subsubsection{Hyper-parameter Analysis}

We use the model's key hyperparameter $\mu$, as described in Algorithm~\ref{cl2_algorithm} and Section~\ref{sec43}, to govern the update frequency of the three CL Schedulers, represented in the code as S\_Size, T\_Size, Q\_Size. These three adjustable hyperparameters show minimal sensitivity to hyperparameter tuning, with results such as RMSE and Q90 for horizon 12 on the PEMS08 datasets shown in~\ref{hyper}.

\input{table/hyper}

\subsubsection{Computational Analysis}
\begin{figure}[h!]
  \centering
  \includegraphics[width=\linewidth]{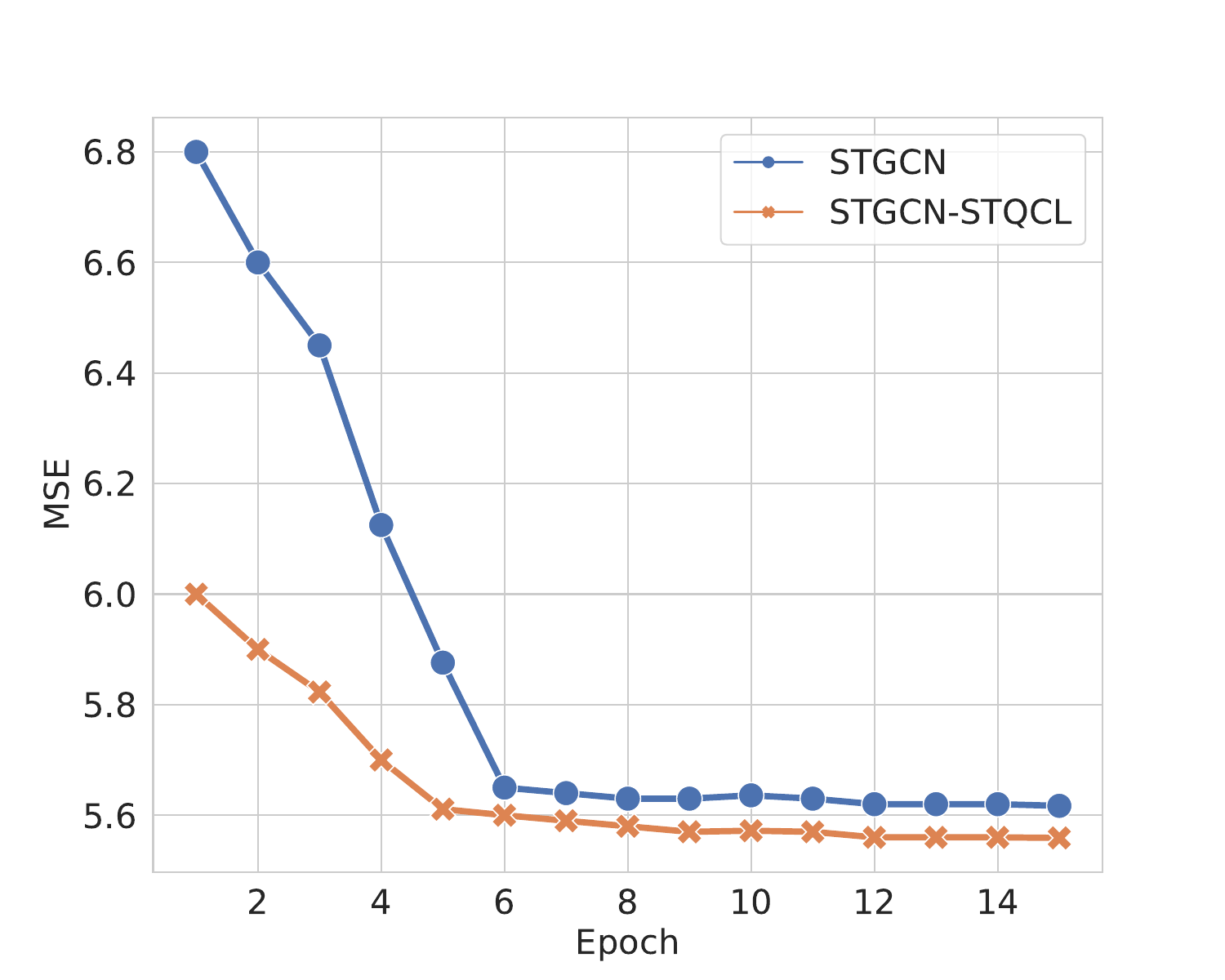}
  \caption{Performance comparison with different step sizes.}
  \label{comp}
\end{figure}

Our model complexity is based on the basic model, and our design shares the structure of the basic model, so it does not increase the graphics card memory usage. On the PEMS04 dataset, both the basic model STGCN and our STGCN-STQCL model have a parameter count of 172,468. Moreover, by removing the predictor of the basic model and employing our stacking model, the graphics card memory usage is not only maintained but also reduced compared to the basic model. For instance, on the PEMS04 dataset, the graphics card memory usage for STGCN and STGCN+STQCL are 3170MB and 2710MB, respectively. However, we acknowledge the increase in operational costs, which is one of our limitations, and we plan to address this issue in the future for improvement. Our framework enables the base model to converge earlier. Taking the PEMS08 data as an example shown in Figure~\ref{comp}, at the third epoch, the test loss of STGCN is 6.35, while for STGCN-STQCL it is 5.823. At the fifth epoch, the test loss for STGCN is 5.668, whereas for STGCN-STQCL it is 5.611. At the tenth epoch, the test loss for STGCN is 5.636, compared to 5.57 for STGCN-STQCL.

\section{Lessons}
In our exploration of integrating CL with spatio-temporal forecasting problem, we have learned several lessons. Therefore, this section will elaborate on a few issues we encountered. To facilitate a clearer understanding, our discussion proceeds with the basic model, STGCN.

\subsection{Lesson 1: How to Start the Curriculum Learning}
\begin{figure}[h!]
  \centering
  \begin{subfigure}[b]{0.495\linewidth}
    \centering
    \includegraphics[width=\linewidth]{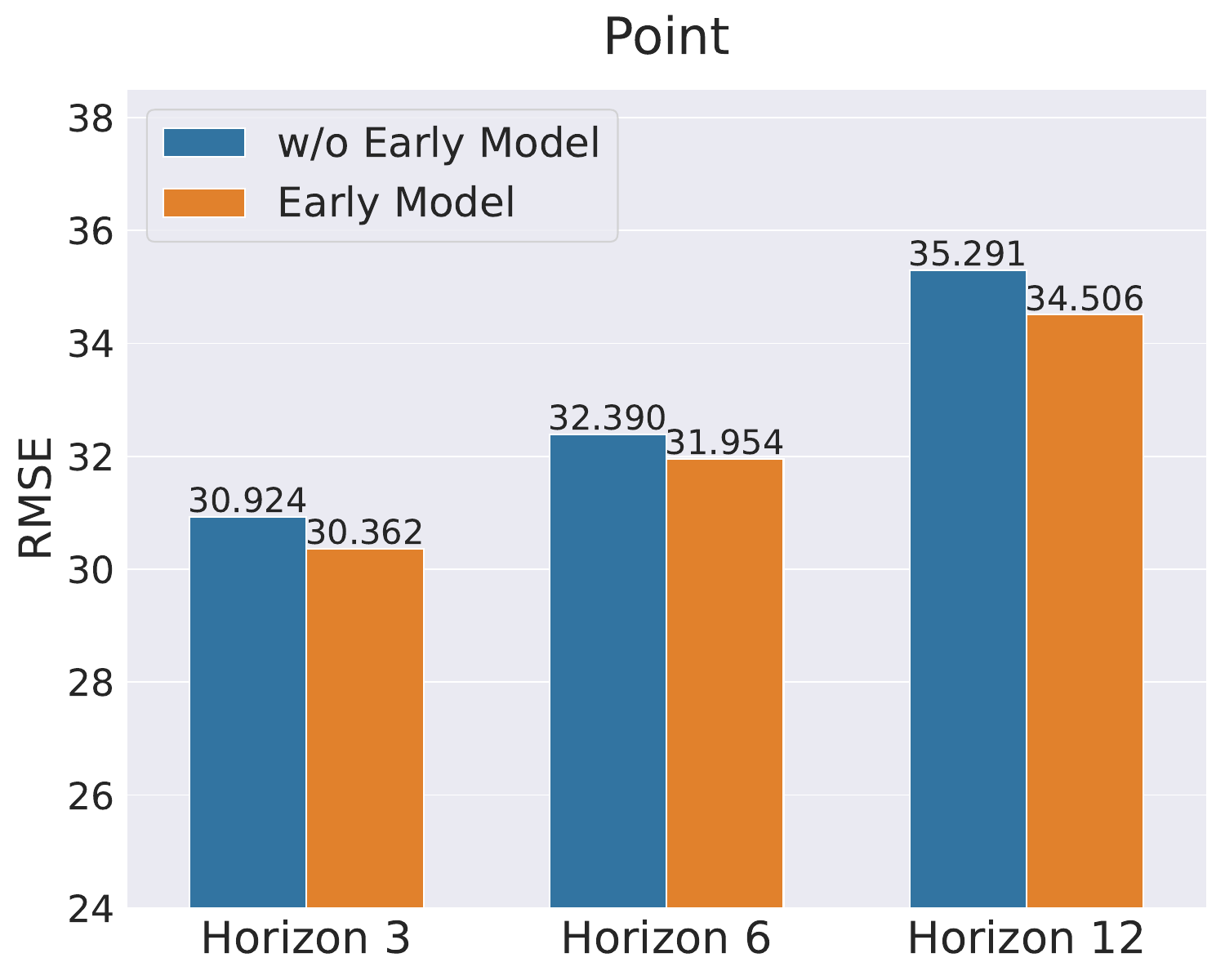}
    \caption{}
    \label{les1:sub1}
  \end{subfigure}
  \hfill
  \begin{subfigure}[b]{0.495\linewidth}
    \centering
    \includegraphics[width=\linewidth]{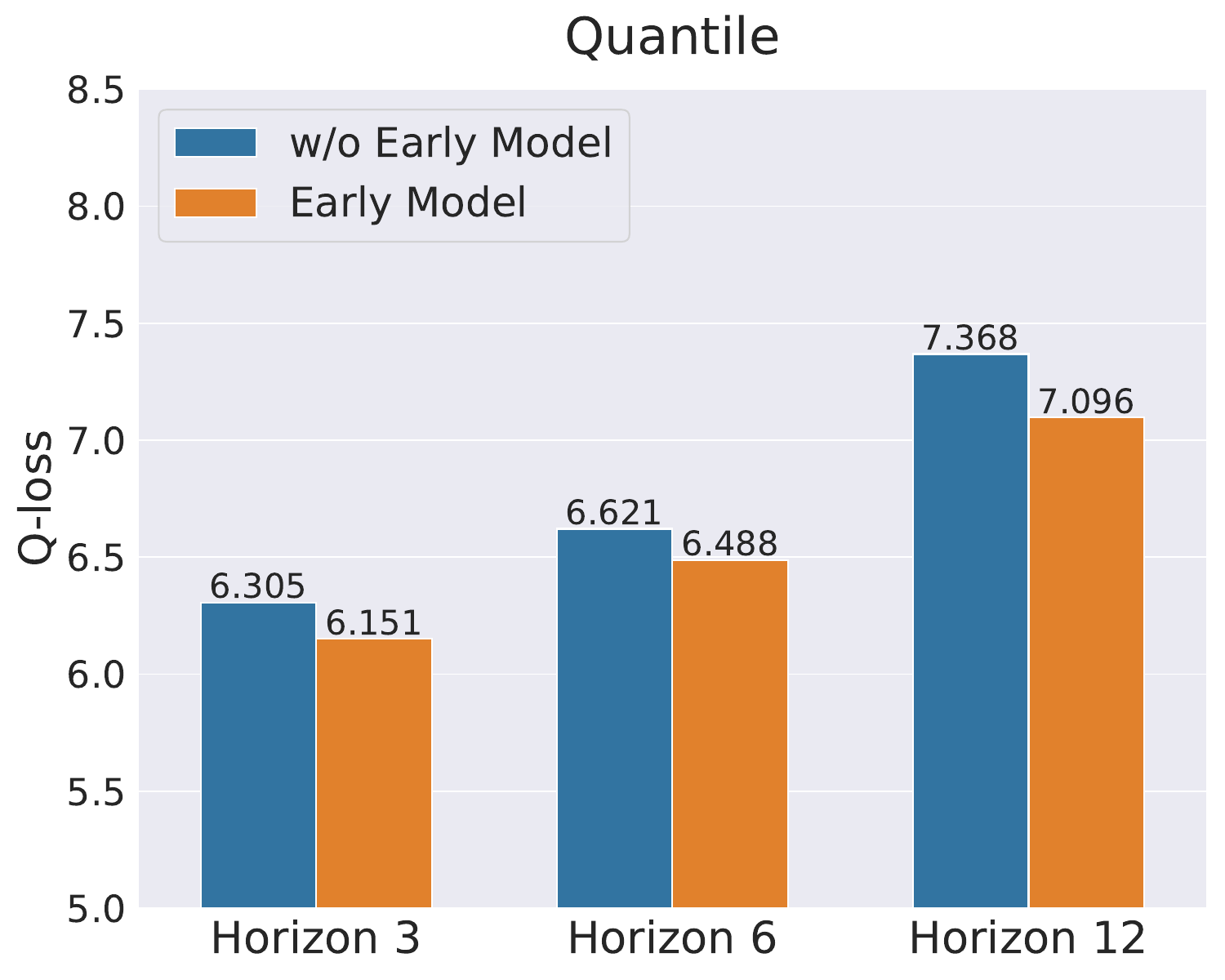}
    \caption{}
    \label{les1:sub2}
  \end{subfigure}
  \caption{Ablation experiments of Early Model. Point predictions are evaluated using RMSE and quantile predictions are gauged with the average of Q10, Q50, and Q90.}
  \label{les1}
\end{figure}

Initially, Starting curriculum learning depends on an initial parameter $\lambda_0$. Therefore, we intuitively believe that guidance from a partially trained model is required, rather than relying solely on the model that's merely evolved from the first set of iteration data to determine difficulty. We have experimented with using an initial model that had been trained for a short period to calculate difficulty, and achieved notable improvements. As illustrated in Figure~\ref{les1}, the performance using an initial model significantly outperforms that without any initial model, demonstrating that a brief initial training period for the basic model is crucial for accurately estimating task difficulty.

\subsection{Lesson 2: Easy to Hard vs. Hard to Easy?}

\begin{figure}[h!]
  \centering
  \begin{subfigure}[b]{0.495\linewidth}
    \centering
    \includegraphics[width=\linewidth]{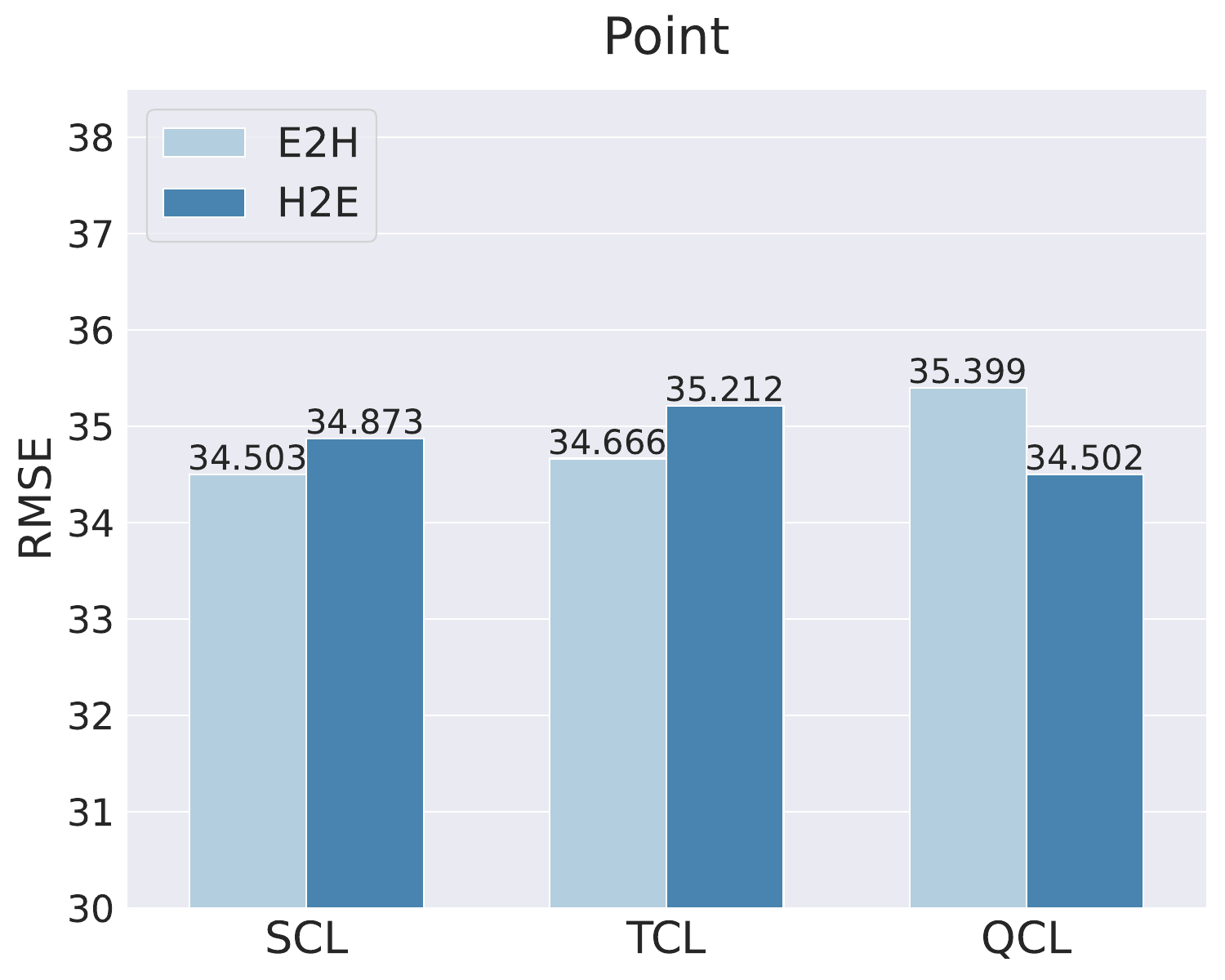}
    \caption{}
    \label{les2:sub1}
  \end{subfigure}
  \hfill
  \begin{subfigure}[b]{0.495\linewidth}
    \centering
    \includegraphics[width=\linewidth]{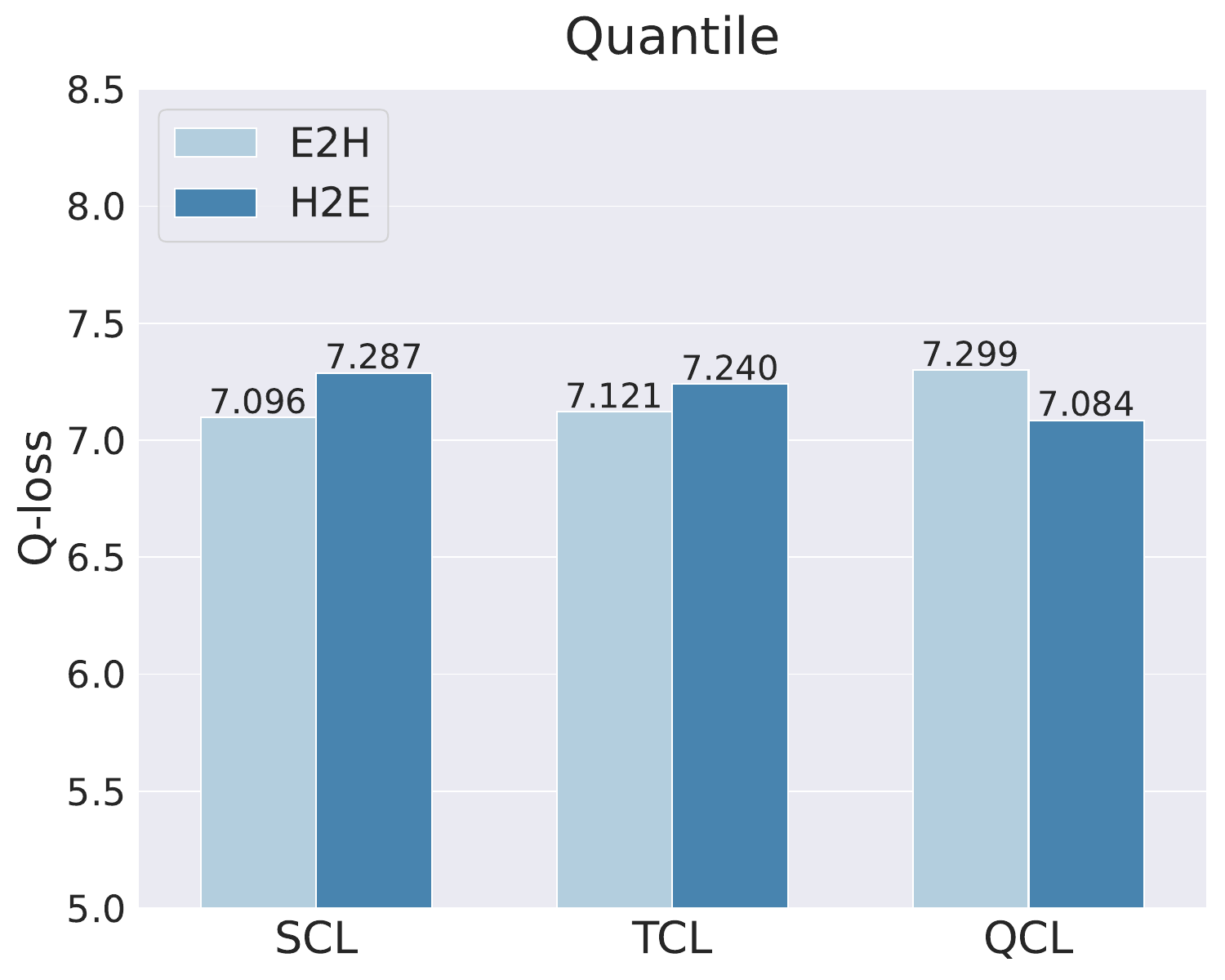}
    \caption{}
    \label{les2:sub2}
  \end{subfigure}
  \caption{The impact of sample difficulty progression on CL scheduler efficacy: Easy-to-Hard vs. Hard-to-Easy.}
  \label{les2}
\end{figure}

In curriculum learning, the progression might start with simple samples and move towards harder ones, or it can begin with hard samples and advance to easier ones. Consequently, in our study, we have explored the sequence of difficulty across three distinct modules: SCL, TCL, and QCL. Through comprehensive experimentation, we have identified the optimal order of complexity for each module. As Figure~\ref{les2} shows, we select RMSE and the average values of Q10, Q50, and Q90, termed Q-Loss for evaluation. SCL and TCL require a data scaling process that moves from easy to hard, whereas QCL necessitates a task sequence from hard to easy. Since SCL and TCL integrate curriculum learning by enhancing data complexity, the gradual increase in difficulty engenders distinct convergence models from both spatial and temporal perspectives. QCL directly modifies the loss function, incorporating curriculum learning at the task level. As illustrated in Figure~\ref{fig2}, transitioning from tasks with lower tolerance levels to those with higher tolerance levels can enhance the model's ability to accurately predict both upper and lower boundaries.

\subsection{Lesson 3: Comparison of Integrated vs. Single CL Model}

\begin{figure}[h!]
  \centering
  \begin{subfigure}[b]{0.85\linewidth}
    \centering
    \includegraphics[width=\linewidth]{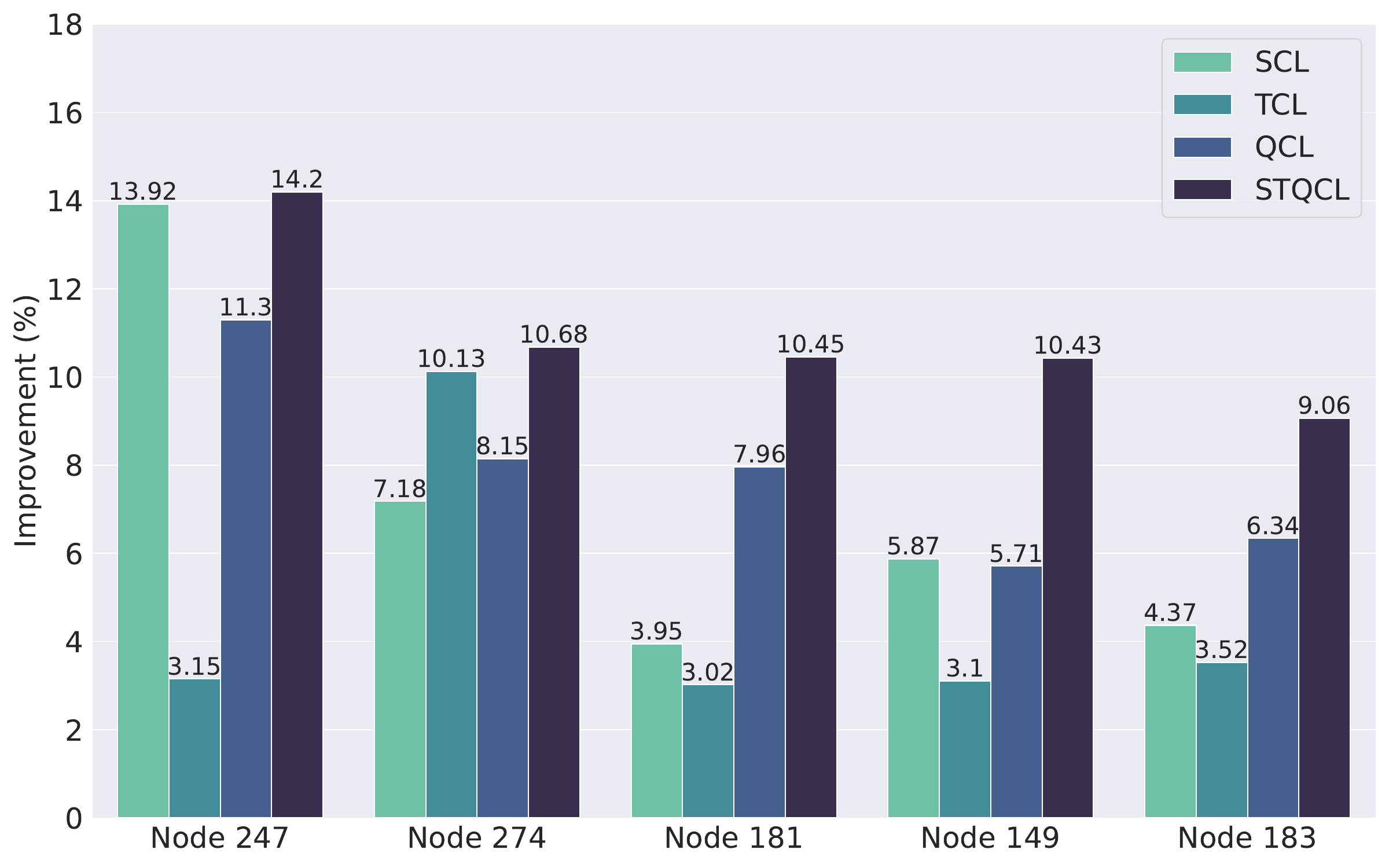}
    \caption{Top 5 performance improvement on RMSE metric in point predictions.}
    \label{les3:sub1}
  \end{subfigure}
  \hfill
  \begin{subfigure}[b]{0.85\linewidth}
    \centering
    \includegraphics[width=\linewidth]{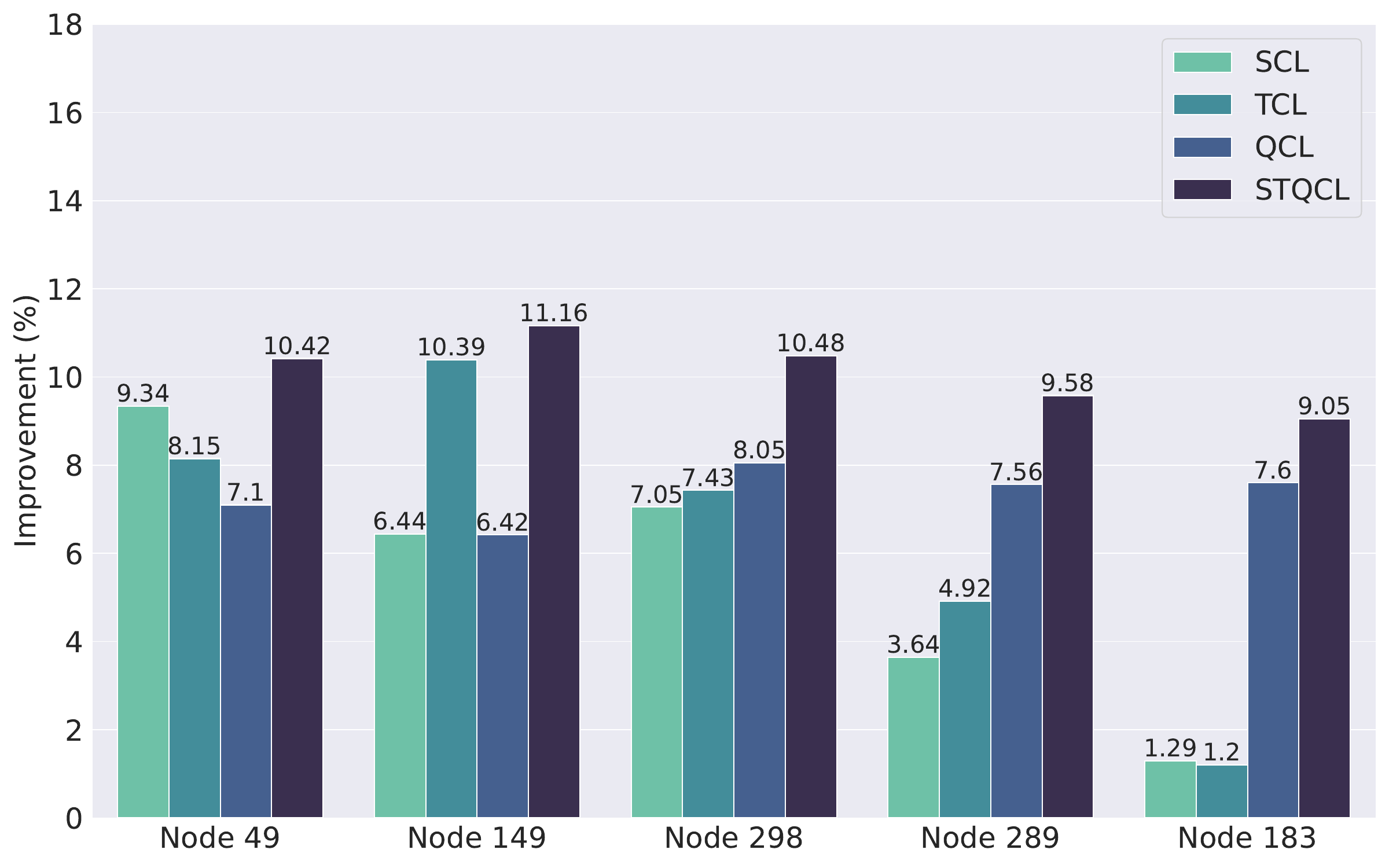}
    \caption{Top 5 performance improvement on Quantile metric in quantile predictions}
    \label{les3:sub2}
  \end{subfigure}
  \caption{Evaluating Integrated vs. Single CL Models: enhancing performance and comprehensiveness.}
  \label{les3}
\end{figure}

Different types of curriculum learning schedulers can assimilate diverse sets of information. Integrating various curriculum modules, however, yields a more comprehensive and unified perspective, culminating in significantly enhanced performance. As illustrated in Figure~\ref{les3}, among the top 5 nodes showing performance improvements, SCL, TCL, and QCL each demonstrate periods of superior performance. The integration of these three types of curriculum learning further enhances these improvements.

\section{Conclusion}
In this research, we introduced an innovative curriculum learning framework specifically designed to enhance spatio-temporal quantile forecasting. By systematically integrating curriculum learning strategies with spatio-temporal forecasting tasks, our framework demonstrates significant improvements. Extensive experimental validation on diverse datasets shows that our approach not only outperforms existing state-of-the-art methods but also offers novel insights into the dynamics of spatio-temporal prediction. 
\subsection{Limitations}
Our framework of the curriculum may not be suitable for multidimensional data, and our framework might become more complex, which could lead to a lack of efficiency for such issues. Additionally, since our framework involves feeding data into three CL Modules, our design ensures that the graphics card memory does not increase, although the runtime will be extended. This limitation can also be addressed in the future by training the same set of data but optimizing different parameters through backpropagation.
\subsection{Future work}
To broaden the utility and impact of our framework, future endeavors will focus on incorporating adaptive curriculum learning mechanisms to dynamically adjust to the complexities of ST data. Additionally, we aim to enhance forecasting richness by integrating multi-modal data sources. Investigating our framework's application in areas such as urban planning and environmental monitoring holds the promise of significant contributions to these fields.



\begin{acks}
We would like to acknowledge the support of Cisco’s National Industry Innovation Network (NIIN) Research Chair Program and the ARC Centre of Excellence for Automated Decision-Making and Society (CE200100005). We acknowledge the resources and services from the National Computational Infrastructure (NCI), which is supported by the Australian Government.
\end{acks}

\bibliographystyle{ACM-Reference-Format}
\bibliography{sample-base}

\clearpage
\appendix
\section{Appendix}
\label{appendix}
\subsection{Results}
As Table~\ref{extra_experiment} shown, our proposed STQCL also enhance the performance in PEMS03, PEMS07, PEMS08, PEMSBAY.
\input{table/appen1}

\end{document}

%% file: table/data.tex
\begin{table}[H]
\caption{Statistics of the datasets in this work.}
\label{data}
\centering

\renewcommand\arraystretch{1.07}
\setlength{\tabcolsep}{1.2mm}{
\begin{tabular}{lccccc}
\hline
\textbf{Dataset}  & Nodes & Start time  & End time    & Samples    &  Interval  \\
\hline
\hline
PEMS03      & 307  & 09/01/2018  & 11/30/2018   &  26,208 & 5min  \\
PEMS04      & 307  & 01/01/2018  & 02/28/2018   &  16,992 & 5min  \\
PEMS07      & 883  & 05/01/2017  & 08/06/2017   &  28,224 & 5min  \\
PEMS08      & 307  & 07/01/2016  & 08/31/2016   &  17,856 & 5min  \\
METR-LA     & 207  & 03/01/2012     & 06/27/2012 & 34,272  & 5min   \\
PEMSBAY    & 325  & 03/01/2017     & 06/30/2017 & 52,116  & 5min   \\
\hline
\end{tabular}}
\end{table}

%% file: table/main_experiment.tex
{
\setlength{\tabcolsep}{2.17pt} 
\begin{table*}[t!]
\caption{Overall comparison.}
\label{main_experiment}
\small
\renewcommand{\arraystretch}{1.17}
\begin{tabular}{lccccccccccccccccc}
\hline


\multicolumn{2}{c}{\multirow{3}{*}{\textbf{Model}}} & \multicolumn{8}{c|}{PEMS04}  & \multicolumn{8}{c}{METR-LA}    \\ \cline{3-18}

\multicolumn{2}{c}{}   & \multicolumn{4}{c}{Base Model} & \multicolumn{4}{c|}{Base Model + STQCL} & \multicolumn{4}{c}{Base Model} & \multicolumn{4}{c}{Base Model + STQCL} \\  \cline{3-18}
\multicolumn{2}{c}{}    & Horizon & 3 & 6 & 12  & Horizon & 3 & 6 & \multicolumn{1}{c|}{12}   & Horizon & 3 & 6 & 12  & Horizon & 3 & 6 & 12   \\ \hline

\multirow{6}{*}{STGCN} & \multirow{3}{*}{Point} & RMSE & 30.522 & 32.305 & 35.098 & RMSE & {\ul \textbf{29.884}} & {\ul \textbf{31.567}} & \multicolumn{1}{c|}{{\ul \textbf{34.177}}}  & RMSE & 6.042 & 7.246 & 8.435   & RMSE & {\ul \textbf{5.806}} & {\ul \textbf{6.919}} & {\ul \textbf{7.974}}        \\

 &   & MAE & 19.167 & 20.234 & 22.192 & MAE & {\ul \textbf{18.689}} & {\ul \textbf{19.683}} & \multicolumn{1}{c|}{{\ul \textbf{21.540}}} & MAE & 3.132 & 3.622 & 4.171   & MAE &  {\ul \textbf{3.027}} & {\ul \textbf{3.488}} & {\ul \textbf{3.999}}    \\
 
  &   & MAPE & 11.350 & 11.914 & 13.113 & MAPE & {\ul \textbf{11.022}} & {\ul \textbf{11.529}} & \multicolumn{1}{c|}{{\ul \textbf{12.586}}} & MAPE & 8.665 & 10.638 & 12.873   & MAPE &  {\ul \textbf{8.249}} & {\ul \textbf{10.086}} & {\ul \textbf{12.095}}        \\
  
& \multirow{3}{*}{Quantile} & Q10 & 4.552 & 4.829 & 5.318 & Q10 & {\ul \textbf{4.419}} & {\ul \textbf{4.696}} & \multicolumn{1}{c|}{{\ul \textbf{5.171}}} & Q10 & 0.930 & 1.113 & 1.367   & Q10 &  {\ul \textbf{0.878}} & {\ul \textbf{1.045}} & {\ul \textbf{1.257}}     \\

 &   & Q50 & 9.702 & 10.243 & 11.229 & Q50 & {\ul \textbf{9.463}} & {\ul \textbf{9.967}} & \multicolumn{1}{c|}{{\ul \textbf{10.906}}} & Q50 & 1.566 & 1.811 & 2.086   & Q50 &  {\ul \textbf{1.514}} & {\ul \textbf{1.744}} & {\ul \textbf{2.000}}     \\
  &   & Q90 & 4.418 & 4.707 & 5.228 & Q90 & {\ul \textbf{4.273}} & {\ul \textbf{4.551}} & \multicolumn{1}{c|}{{\ul \textbf{5.039}}} & Q90 & 0.607 & 0.669 & 0.736   & Q90 &  {\ul \textbf{0.586}} & {\ul \textbf{0.648}} & {\ul \textbf{0.720}}     \\ \hline

\multirow{6}{*}{DCRNN} & \multirow{3}{*}{Point} & RMSE & 30.181 & 32.280 & 36.133& RMSE & {\ul \textbf{29.954}} & {\ul \textbf{31.894}} & \multicolumn{1}{c|}{{\ul \textbf{35.657}}}  & RMSE & 5.370 & 6.415 & 7.578   & RMSE & {\ul \textbf{5.167}} & {\ul \textbf{6.304}} & {\ul \textbf{7.495}}        \\

 &   & MAE & 19.193 & 20.608 & 23.419   & MAE &  {\ul \textbf{19.001}} & {\ul \textbf{20.154}} & {\ul \textbf{22.947}} & MAE & 2.825 & 3.232 & 3.760 & MAE & {\ul \textbf{2.684}} & {\ul \textbf{3.169}} & \multicolumn{1}{c}{{\ul \textbf{3.684}}}    \\
 
  &   & MAPE & 11.481 & 12.316 & 14.144 & MAPE & {\ul \textbf{11.237}} & {\ul \textbf{12.125}} & \multicolumn{1}{c|}{{\ul \textbf{13.984}}} & MAPE & 7.356 & 8.869 & 10.828   & MAPE &  {\ul \textbf{7.231}} & {\ul \textbf{8.795}} & {\ul \textbf{10.573}}        \\
  
& \multirow{3}{*}{Quantile} & Q10 & 4.554 & 4.923 & 5.646 & Q10 & {\ul \textbf{4.416}} & {\ul \textbf{4.785}} & \multicolumn{1}{c|}{{\ul \textbf{5.462}}} & Q10 & 0.812 & 0.951 & 1.133   & Q10 &  {\ul \textbf{0.783}} & {\ul \textbf{0.894}} & {\ul \textbf{1.053}}     \\

 &   & Q50 & 9.722 & 10.438 & 11.858 & Q50 & {\ul \textbf{9.651}} & {\ul \textbf{10.438}} & \multicolumn{1}{c|}{{\ul \textbf{11.858}}} & Q50 & 1.413 & 1.616 & 1.880  & Q50 &  {\ul \textbf{1.364}} & {\ul \textbf{1.514}} & {\ul \textbf{1.801}}     \\
  &   & Q90 & 4.437 & 4.823 & 5.639 & Q90 & {\ul \textbf{4.341}} & {\ul \textbf{4.574}} & \multicolumn{1}{c|}{{\ul \textbf{5.484}}} & Q90 & 0.571 & 0.634 & 0.715   & Q90 &  {\ul \textbf{0.506}} & {\ul \textbf{0.576}} & {\ul \textbf{0.649}}     \\ \hline    
  
\multirow{6}{*}{GWN} & \multirow{3}{*}{Point} & RMSE & 29.237 & 30.852 & 33.124 & RMSE & {\ul \textbf{28.587}} & {\ul \textbf{30.038}} & \multicolumn{1}{c|}{{\ul \textbf{31.910}}}  & RMSE & 5.420 & 6.398 & 7.420   & RMSE & {\ul \textbf{5.281}} & {\ul \textbf{6.187}} & {\ul \textbf{7.144}}        \\

 &   & MAE & 18.050 & 19.020 & 20.579 & MAE & {\ul \textbf{17.625}} & {\ul \textbf{18.490}} & \multicolumn{1}{c|}{{\ul \textbf{19.820}}} & MAE & 2.805 & 3.185 & 3.637   & MAE &  {\ul \textbf{2.740}} & {\ul \textbf{3.096}} & {\ul \textbf{3.510}}    \\
 
  &   & MAPE & 10.805 & 11.366 & 12.325 & MAPE & {\ul \textbf{10.496}} & {\ul \textbf{10.982}} & \multicolumn{1}{c|}{{\ul \textbf{11.780}}} & MAPE & 7.332 & 8.771 & 10.449   & MAPE &  {\ul \textbf{7.149}} & {\ul \textbf{8.519}} & {\ul \textbf{10.023}}        \\
  
& \multirow{3}{*}{Quantile} & Q10 & 4.313 & 4.621 & 5.059 & Q10 & {\ul \textbf{4.186}} & {\ul \textbf{4.458}} & \multicolumn{1}{c|}{{\ul \textbf{4.812}}} & Q10 & 0.803 & 0.929 & 1.081   & Q10 &  {\ul \textbf{0.781}} & {\ul \textbf{0.899}} & {\ul \textbf{1.038}}     \\

 &   & Q50 & 9.140 & 9.634 & 10.416 & Q50 & {\ul \textbf{8.927}} & {\ul \textbf{9.365}} & \multicolumn{1}{c|}{{\ul \textbf{10.037}}} & Q50 & 1.402 & 1.593 & 1.818   & Q50 &  {\ul \textbf{1.370}} & {\ul \textbf{1.548}} & {\ul \textbf{1.755}}     \\
  &   & Q90 & 4.105 & 4.378 & 4.794 & Q90 & {\ul \textbf{4.002}} & {\ul \textbf{4.248}} & \multicolumn{1}{c|}{{\ul \textbf{4.620}}} & Q90 & 0.562 & 0.621 & 0.691   & Q90 &  {\ul \textbf{0.544}} & {\ul \textbf{0.598}} & {\ul \textbf{0.666}}     \\ \hline

\multirow{6}{*}{STNorm} & \multirow{3}{*}{Point} & RMSE & 29.551 & 31.281 & 33.476 & RMSE & {\ul \textbf{29.263}} & {\ul \textbf{30.867}} & \multicolumn{1}{c|}{{\ul \textbf{32.839}}}  & RMSE & 5.475 & 6.434 & 7.443   & RMSE & {\ul \textbf{5.376}} & {\ul \textbf{6.296}} & {\ul \textbf{7.278}}        \\
 &   & MAE & 18.222 & 19.197 & 20.756 & MAE & {\ul \textbf{18.166}} & {\ul \textbf{19.085}} & \multicolumn{1}{c|}{{\ul \textbf{20.395}}} & MAE & 2.826 & 3.192 & 3.627   & MAE &  {\ul \textbf{2.773}} & {\ul \textbf{3.134}} & {\ul \textbf{3.566}}    \\
 &   & MAPE & 10.831 & 11.359 & 12.428 & MAPE & {\ul \textbf{10.806}} & {\ul \textbf{11.314}} & \multicolumn{1}{c|}{{\ul \textbf{12.131}}} & MAPE & 7.359 & 8.799 & 10.500   & MAPE &  {\ul \textbf{7.278}} & {\ul \textbf{8.688}} & {\ul \textbf{10.392}}        \\
& \multirow{3}{*}{Quantile} & Q10 & 4.308 & 4.637 & 5.090 & Q10 & {\ul \textbf{4.258}} & {\ul \textbf{4.550}} & \multicolumn{1}{c|}{{\ul \textbf{4.939}}} & Q10 & 0.808 & 0.934 & 1.085   & Q10 &  {\ul \textbf{0.796}} & {\ul \textbf{0.920}} & {\ul \textbf{1.069}}     \\
 &   & Q50 & 9.226 & 9.722 & 10.506 & Q50 & {\ul \textbf{9.199}} & {\ul \textbf{9.665}} & \multicolumn{1}{c|}{{\ul \textbf{10.326}}} & Q50 & 1.413 & 1.596 & 1.814   & Q50 &  {\ul \textbf{1.387}} & {\ul \textbf{1.567}} & {\ul \textbf{1.783}}     \\
 &   & Q90 & 4.146 & 5.090 & 4.825 & Q90 & {\ul \textbf{4.116}} & {\ul \textbf{4.365}} & \multicolumn{1}{c|}{{\ul \textbf{4.741}}} & Q90 & 0.561 & 0.614 & 0.683   & Q90 &  {\ul \textbf{0.550}} & {\ul \textbf{0.601}} & {\ul \textbf{0.670}}     \\ \hline

\multirow{6}{*}{SCINet} & \multirow{3}{*}{Point} & RMSE & 29.644 & 30.923 & 33.605 & RMSE & {\ul \textbf{29.571}} & {\ul \textbf{30.871}} & \multicolumn{1}{c|}{{\ul \textbf{33.371}}}  & RMSE & 5.902 & 6.713 & 7.794   & RMSE & {\ul \textbf{5.814}} & {\ul \textbf{6.637}} & {\ul \textbf{7.687}}        \\
 &   & MAE & 18.705 & 19.531 & 21.550 & MAE & {\ul \textbf{18.588}} & {\ul \textbf{19.421}} & \multicolumn{1}{c|}{{\ul \textbf{21.229}}} & MAE & 3.278 & 3.671 & 4.221   & MAE &  {\ul \textbf{3.216}} & {\ul \textbf{3.609}} & {\ul \textbf{4.147}}    \\
 &   & MAPE & 11.163 & 11.666 & 13.024 & MAPE & {\ul \textbf{11.042}} & {\ul \textbf{11.499}} & \multicolumn{1}{c|}{{\ul \textbf{12.618}}} & MAPE & 8.940 & 10.374 & 12.305   & MAPE &  {\ul \textbf{8.664}} & {\ul \textbf{10.154}} & {\ul \textbf{12.042}}        \\
& \multirow{3}{*}{Quantile} & Q10 & 4.625 & 4.829 & 5.295 & Q10 & {\ul \textbf{4.579}} & {\ul \textbf{4.807}} & \multicolumn{1}{c|}{{\ul \textbf{5.236}}} & Q10 & 0.977 & 1.112 & 1.312   & Q10 &  {\ul \textbf{0.967}} & {\ul \textbf{1.097}} & {\ul \textbf{1.285}}     \\
 &   & Q50 & 9.471 & 9.883 & 10.899 & Q50 & {\ul \textbf{9.408}} & {\ul \textbf{9.829}} & \multicolumn{1}{c|}{{\ul \textbf{10.741}}} & Q50 & 1.639 & 1.836 & 2.111   & Q50 &  {\ul \textbf{1.608}} & {\ul \textbf{1.804}} & {\ul \textbf{2.073}}     \\
 &   & Q90 & 4.344 & 4.590 & 5.151 & Q90 & {\ul \textbf{4.271}} & {\ul \textbf{4.511}} & \multicolumn{1}{c|}{{\ul \textbf{4.999}}} & Q90 & 0.708 & 0.761 & 0.872   & Q90 &  {\ul \textbf{0.679}} & {\ul \textbf{0.728}} & {\ul \textbf{0.823}}     \\ \hline

\multirow{6}{*}{STAEFormer} & \multirow{3}{*}{Point} & RMSE & 29.292 & 30.931 & 32.922 & RMSE & {\ul \textbf{28.685}} & {\ul \textbf{30.118}} & \multicolumn{1}{c|}{{\ul \textbf{31.828}}}  & RMSE & 5.214 & 6.158 & 7.153   & RMSE & {\ul \textbf{5.038}} & {\ul \textbf{5.948}} & {\ul \textbf{6.941}}        \\
 &   & MAE & 17.589 & 18.371 & 19.543 & MAE & {\ul \textbf{17.384}} & {\ul \textbf{18.088}} & \multicolumn{1}{c|}{{\ul \textbf{19.137}}} & MAE & 2.707 & 3.042 & 3.430   & MAE &  {\ul \textbf{2.629}} & {\ul \textbf{2.956}} & {\ul \textbf{3.349}}    \\
 &   & MAPE & 11.993 & 12.435 & 13.171 & MAPE & {\ul \textbf{10.456}} & {\ul \textbf{10.796}} & \multicolumn{1}{c|}{{\ul \textbf{11.370}}} & MAPE & 7.018 & 8.353 & 10.047   & MAPE &  {\ul \textbf{6.789}} & {\ul \textbf{8.100}} & {\ul \textbf{9.703}}        \\
& \multirow{3}{*}{Quantile} & Q10 & 4.190 & 4.486 & 4.844 & Q10 & {\ul \textbf{4.150}} & {\ul \textbf{4.412}} & \multicolumn{1}{c|}{{\ul \textbf{4.726}}} & Q10 & 0.765 & 0.886 & 1.036   & Q10 &  {\ul \textbf{0.743}} & {\ul \textbf{0.864}} & {\ul \textbf{1.010}}     \\
 &   & Q50 & 8.794 & 9.186 & 9.772 & Q50 & {\ul \textbf{8.808}} & {\ul \textbf{9.163}} & \multicolumn{1}{c|}{{\ul \textbf{9.693}}} & Q50 & 1.353 & 1.521 & 1.717   & Q50 &  {\ul \textbf{1.315}} & {\ul \textbf{1.478}} & {\ul \textbf{1.675}}     \\
 &   & Q90 & 3.921 & 4.141 & 4.482 & Q90 & {\ul \textbf{3.942}} & {\ul \textbf{4.148}} & \multicolumn{1}{c|}{{\ul \textbf{4.449}}} & Q90 & 0.538 & 0.588 & 0.647   & Q90 &  {\ul \textbf{0.524}} & {\ul \textbf{0.575}} & {\ul \textbf{0.637}}     \\ \hline

\end{tabular}
\end{table*}
}

%% file: table/point.tex
{
\setlength{\tabcolsep}{2.28pt} 

\begin{table}[]
\caption{Comparison of point forecasting from basic models and STQCL enhanced models.}
\label{point}
\small
\renewcommand{\arraystretch}{1.28}
\begin{tabular}{cccccccc}
\hline
\multicolumn{2}{c}{\multirow{3}{*}{model}} & \multicolumn{6}{c}{PEMS04}                                             \\  \cline{3-8}
\multicolumn{2}{c}{}   & \multicolumn{2}{c}{Horizon 3} & \multicolumn{2}{c}{Horizon 6} & \multicolumn{2}{c}{Horizon 12} \\ \cline{3-8}
\multicolumn{2}{c}{}                       & Point     & STQCL     & Point     & STQCL     & Point      & STQCL     \\  \hline{}
\multirow{3}{*}{STGCN}        & RMSE       &    30.826       &     {\ul \textbf{29.884}}      &  32.246         &   {\ul \textbf{31.567}}         &       34.699     &      {\ul \textbf{34.177}}     \\
& MAE  & 19.522 & {\ul \textbf{18.689}}  & 20.375  &  {\ul \textbf{19.683}} &  22.236 & {\ul \textbf{21.540}}  \\
& MAPE & 11.649  & {\ul \textbf{11.022}} &  12.121 &  {\ul \textbf{11.529}} & 13.318 & {\ul \textbf{12.586}}     \\  \hline

\multirow{3}{*}{DCRNN}        & RMSE       &    30.079       &     {\ul \textbf{29.954}}      &  32.134        &   {\ul \textbf{31.894}}         &       35.893     &      {\ul \textbf{35.657}}     \\
& MAE  & 19.138 & {\ul \textbf{19.001}}  & 20.329 &  {\ul \textbf{20.154}} &  23.146 & {\ul \textbf{22.947}}  \\
& MAPE & 11.649  & {\ul \textbf{11.237}} &  12.121 &  {\ul \textbf{12.125}} & 14.042 & {\ul \textbf{13.984}}     \\  \hline

\multirow{3}{*}{GWN}        & RMSE       &    28.771       &     {\ul \textbf{28.587}}      &  30.322        &   {\ul \textbf{30.038}}         &       32.285     &      {\ul \textbf{31.910}}     \\
& MAE  & 17.720 & {\ul \textbf{17.625}}  & 18.579  &  {\ul \textbf{18.490}} &  19.942 & {\ul \textbf{19.820}}  \\
& MAPE & 10.500  & {\ul \textbf{10.496}} &  11.017 &  {\ul \textbf{10.982}} & 11.837 & {\ul \textbf{11.780}}     \\  \hline     

\multirow{3}{*}{STNorm}        & RMSE       &   29.734      &     {\ul \textbf{29.263}}      &  31.458       &   {\ul \textbf{30.867}}         &     33.841     &      {\ul \textbf{32.839}}     \\
& MAE  & 18.673 & {\ul \textbf{18.166}}  & 19.705 &  {\ul \textbf{19.085}} &  21.442 & {\ul \textbf{20.395}}  \\
& MAPE & 11.173  & {\ul \textbf{10.806}} &  11.827 &  {\ul \textbf{11.314}} & 13.002 & {\ul \textbf{12.131}}     \\ \hline 

\multirow{3}{*}{SCINet}        & RMSE       &    29.981       &     {\ul \textbf{29.571}}      &  31.319         &   {\ul \textbf{30.871}}         &       34.161    &      {\ul \textbf{33.371}}     \\
& MAE  & 19.140 & {\ul \textbf{18.588}}  & 19.997  &  {\ul \textbf{19.421}} &  22.196 & {\ul \textbf{21.229}}  \\
& MAPE & 11.652  & {\ul \textbf{11.042}} &  12.104 &  {\ul \textbf{11.499}} & 13.677 & {\ul \textbf{12.618}}     \\  \hline

\multirow{3}{*}{STAEFormer}        & RMSE       &    28.773       &     {\ul \textbf{28.685}}      &  30.178        &   {\ul \textbf{30.118}}         &       31.902     &      {\ul \textbf{31.828}}     \\
& MAE  & 17.466 & {\ul \textbf{17.384}}  & 18.206  &  {\ul \textbf{18.088}} &  19.268 & {\ul \textbf{19.137}}  \\
& MAPE & 11.614  & {\ul \textbf{10.496}} &  12.025 &  {\ul \textbf{10.982}} & 12.728 & {\ul \textbf{11.780}}     \\  \hline     

\end{tabular}
\end{table}
}

%% file: table/quantile.tex
{
\setlength{\tabcolsep}{0.7pt} 
\begin{table}[]

\small
\caption{Comparison of our modified quantile models and the tradictional quantile models.}
\label{quantile_com}
\renewcommand{\arraystretch}{1.29}
\begin{tabular}{lccccccccc}
\hline
\multirow{3}{*}{Model} & \multicolumn{9}{c}{PEMS04}                                                                     \\ \cline{2-10}
                       & \multicolumn{3}{c}{Horizon 3} & \multicolumn{3}{c}{Horizon 6} & \multicolumn{3}{c}{Horizon 12} \\ \cline{2-10}
                       & Q10      & Q50      & Q90     & Q10      & Q50      & Q90     & Q10      & Q50      & Q90      \\ \hline
DeepAR                 &  5.004        & 10.688        & 5.076       & 5.942        & 12.064        & 6.245       & 6.753        & 12.668        & 8.129       \\ \hline
MQRNN                  &    4.521      &     9.763     &   4.581      &    5.073      &     10.816     &     5.018    &    6.089      &    12.706      &   5.745       \\ \hline
STGCN-Quantile         &    4.552 & 9.702    &   4.418      &    4.829      &    10.243      &  4.707       &    5.318      &     11.229     &   5.228       \\ \cline{2-10}
STGCN-STQCL            &   4.419       &    9.463      &   4.273      &    4.696      &   9.967       &    4.551     &   5.171       &     10.906     &    5.039      \\ \hline
STAEF-Quantile         &    {\ul 4.190}       &   {\ul 8.794}       &        {\ul 3.921} &     {\ul 4.486}     &    {\ul 9.186}      &     {\ul 4.141}    &     {\ul 4.844}     &   {\ul 9.772}       &      {\ul 4.482}    \\ \cline{2-10}
STAEF-STQCL            &      {\ul \textbf{4.150}}    &    {\ul \textbf{8.808}}      &      {\ul \textbf{3.942}}   &     {\ul \textbf{4.412}}     &    {\ul \textbf{9.163}}      &      {\ul \textbf{4.148}}   &      {\ul \textbf{4.726}}    &      {\ul \textbf{9.693}}    &   {\ul \textbf{4.449}}      \\ \hline
\end{tabular}
\end{table}
}

%% file: table/ablation.tex
\begin{table}[h!]
\caption{Ablation study of three curriculum learning modules.}
\label{ablation}
\centering

\normalsize
\renewcommand\arraystretch{1.17}
\setlength{\tabcolsep}{0.9mm}{
\begin{tabular}{l|cccccc}
\hline
\textbf{Modules}  & RMSE & MAE  & MAPE    & Q10  & Q50  & Q90 \\
\hline
\hline
STGCN-Point & 34.699  & 22.236 & 13.318 & \textbackslash & \textbackslash & \textbackslash  \\
STGCN-Quantile & 35.098  & 22.192 & 13.113 & 5.318 & 11.229 & 5.228  \\
STGCN-SCL    & 34.506 &	21.734 & 12.895	& 5.199 & 10.997 & 5.093  \\
STGCN-TCL    & 34.666 &	21.752 & 12.811 & 5.242 & 11.009 & 5.111  \\
STGCN-QCL    & 34.502  & 21.731  & 12.887 &  5.186  &  10.984 &  5.081 \\
 
STGCN-STQCL  & {\ul \textbf{34.177}} & {\ul \textbf{21.540}} & {\ul \textbf{12.586}}  & {\ul \textbf{5.171}} &  {\ul \textbf{10.906}}   & {\ul \textbf{5.039}} \\

\hline
\end{tabular}}
\end{table}

%% file: table/hyper.tex
\begin{table}[h!]
\caption{Performance comparison with different step sizes.}
\label{hyper}
\centering

\normalsize
\renewcommand\arraystretch{1.07}
\setlength{\tabcolsep}{1.2mm}{
\begin{tabular}{l|l|l|l|l}
\hline
\textbf{S\_Size} & \textbf{T\_Size} & \textbf{Q\_Size} & \textbf{RMSE} & \textbf{Q90} \\
\hline
300 & 300 & 300 & 29.197 & 4.706 \\
700 & 300 & 300 & 29.216 & 4.682 \\
1500 & 300 & 300 & 29.117 & 4.719 \\
3000 & 300 & 300 & 29.090 & 4.697 \\
300 & 700 & 300 & 29.213 & 4.758 \\
300 & 1500 & 300 & 29.167 & 4.685 \\
300 & 3000 & 300 & 29.139 & 4.674 \\
300 & 300 & 700 & 29.233 & 4.758 \\
300 & 300 & 1500 & 29.094 & 4.714 \\
300 & 300 & 3000 & 29.247 & 4.678 \\
\hline
\end{tabular}}
\end{table}

%% file: table/appen1.tex
\begin{table*}[t!]
\centering

\small
\caption{Overall comparison in data PEMS03, PEMS07, PEMS08, PEMSBAY.}
\renewcommand\arraystretch{1.27}
\setlength{\tabcolsep}{0.7mm}{\begin{tabular}{lccccccccccccccccc}
\hline
\multicolumn{2}{c}{\multirow{3}{*}{\textbf{Dataset}}} & \multicolumn{8}{c|}{STGCN}  & \multicolumn{8}{c}{STAEFormer}    \\ \cline{3-18}

\multicolumn{2}{c}{}   & \multicolumn{4}{c}{Base Model} & \multicolumn{4}{c|}{Base Model + STQCL} & \multicolumn{4}{c}{Base Model} & \multicolumn{4}{c}{Base Model + STQCL} \\  \cline{3-18}
\multicolumn{2}{c}{}    & Horizon & 3 & 6 & 12  & Horizon & 3 & 6 & \multicolumn{1}{c|}{12}   & Horizon & 3 & 6 & 12  & Horizon & 3 & 6 & 12   \\ \hline

\multirow{6}{*}{PEMS03} & \multirow{3}{*}{Point} & RMSE & 25.973 & 28.477 & 31.906 & RMSE & \textbf{\underline{25.837}} & \textbf{\underline{27.948}} & \multicolumn{1}{c|}{\textbf{\underline{28.861}}}  & RMSE & 23.502 & 25.803 & 28.861   & RMSE & \textbf{\underline{23.141}} & \textbf{\underline{25.511}} & \textbf{\underline{28.529}}        \\

 &   & MAE & 14.839 & 16.146 & 18.429 & MAE & \textbf{\underline{14.623}} & \textbf{\underline{15.915}} & \multicolumn{1}{c|}{\textbf{\underline{18.069}}} & MAE & 14.327 & 15.510 & 17.287   & MAE & \textbf{\underline{14.028}} & \textbf{\underline{15.241}} & \textbf{\underline{16.973}}    \\
 
  &   & MAPE & 15.468 & 16.437 & 18.687 & MAPE & \textbf{\underline{14.817}} & \textbf{\underline{15.997}} & \multicolumn{1}{c|}{\textbf{\underline{18.436}}} & MAPE & 14.812 & 15.722 & 17.105   & MAPE & \textbf{\underline{14.346}} & \textbf{\underline{15.412}} & \textbf{\underline{16.975}}        \\
  
& \multirow{3}{*}{Quantile} & Q10 & 3.450 & 3.805 & 4.427 & Q10 & \textbf{\underline{3.393}} & \textbf{\underline{3.762}} & \multicolumn{1}{c|}{\textbf{\underline{4.399}}} & Q10 & 3.443 & 3.912 & 4.566   & Q10 & \textbf{\underline{3.381}} & \textbf{\underline{3.844}} & \textbf{\underline{4.475}}     \\

 &   & Q50 & 7.420 & 8.073 & 9.215 & Q50 & \textbf{\underline{7.311}} & \textbf{\underline{7.958}} & \multicolumn{1}{c|}{\textbf{\underline{9.035}}} & Q50 & 7.164 & 7.755 & 8.644   & Q50 & \textbf{\underline{7.014}} & \textbf{\underline{7.621}} & \textbf{\underline{8.486}}     \\
  &   & Q90 & 3.657 & 4.048 & 4.763 & Q90 & \textbf{\underline{3.547}} & \textbf{\underline{3.918}} & \multicolumn{1}{c|}{\textbf{\underline{4.512}}} & Q90 & 3.335 & 3.661 & 4.098   & Q90 & \textbf{\underline{3.262}} & \textbf{\underline{3.582}} & \textbf{\underline{4.026}}     \\ \hline

\multirow{6}{*}{PEMS07} & \multirow{3}{*}{Point} & RMSE & 33.334 & 36.538 & 41.745& RMSE & \textbf{\underline{32.400}} & \textbf{\underline{35.366}} & \multicolumn{1}{c|}{\textbf{\underline{40.057}}}  & RMSE & 30.984 & 33.837 & 37.618   & RMSE & \textbf{\underline{30.349}} & \textbf{\underline{33.178}} & \textbf{\underline{36.735}}        \\

 &   & MAE & 21.203 & 23.239 & 26.715   & MAE & \textbf{\underline{20.472}} & \textbf{\underline{22.257}} & \textbf{\underline{25.377}} & MAE & 18.731 & 20.091 & 22.199 & MAE & \textbf{\underline{18.042}} & \textbf{\underline{19.401}} & \multicolumn{1}{c}{\textbf{\underline{21.271}}}    \\
 
  &   & MAPE & 9.214 & 10.215 & 11.815 & MAPE & \textbf{\underline{8.987}} & \textbf{\underline{9.695}} & \multicolumn{1}{c|}{\textbf{\underline{11.202}}} & MAPE & 8.665 & 10.638 & 12.873   & MAPE & \textbf{\underline{8.249}} & \textbf{\underline{10.086}} & \textbf{\underline{12.095}}        \\
  
& \multirow{3}{*}{Quantile} & Q10 & 5.068 & 5.607 & 6.476 & Q10 & \textbf{\underline{4.923}} & \textbf{\underline{5.397}} & \multicolumn{1}{c|}{\textbf{\underline{6.194}}} & Q10 & 7.921 & 8.472 & 9.513   & Q10 & \textbf{\underline{7.579}} & \textbf{\underline{8.110}} & \textbf{\underline{8.951}}     \\

 &   & Q50 & 10.602 & 11.619 & 13.358 & Q50 & \textbf{\underline{10.236}} & \textbf{\underline{11.129}} & \multicolumn{1}{c|}{\textbf{\underline{12.688}}} & Q50 & 9.366 & 10.046 & 11.099   & Q50 & \textbf{\underline{9.021}} & \textbf{\underline{9.701}} & \textbf{\underline{10.636}}     \\
  &   & Q90 & 5.149 & 5.635 & 6.656 & Q90 & \textbf{\underline{4.838}} & \textbf{\underline{5.319}} & \multicolumn{1}{c|}{\textbf{\underline{6.180}}} & Q90 & 4.365 & 4.712 & 5.274   & Q90 & \textbf{\underline{4.179}} & \textbf{\underline{4.524}} & \textbf{\underline{5.025}}     \\ \hline    
  
\multirow{6}{*}{PEMS08} & \multirow{3}{*}{Point} & RMSE & 23.978 & 26.366 & 30.449 & RMSE & \textbf{\underline{23.490}} & \textbf{\underline{25.761}} & \multicolumn{1}{c|}{\textbf{\underline{29.197}}}  & RMSE & 22.371 & 24.360 & 27.025   & RMSE & \textbf{\underline{21.696}} & \textbf{\underline{23.748}} & \textbf{\underline{26.193}}        \\

 &   & MAE & 15.616 & 16.997 & 19.725 & MAE & \textbf{\underline{15.182}} & \textbf{\underline{16.457}} & \multicolumn{1}{c|}{\textbf{\underline{18.806}}} & MAE & 13.577 & 14.415 & 15.927   & MAE & \textbf{\underline{12.933}} & \textbf{\underline{13.864}} & \textbf{\underline{15.230}}    \\
 
  &   & MAPE & 10.015 & 10.957 & 12.614 & MAPE & \textbf{\underline{9.810}} & \textbf{\underline{10.574}} & \multicolumn{1}{c|}{\textbf{\underline{12.138}}} & MAPE & 8.665 & 10.638 & 12.873   & MAPE & \textbf{\underline{8.249}} & \textbf{\underline{10.086}} & \textbf{\underline{12.095}}        \\
  
& \multirow{3}{*}{Quantile} & Q10 & 3.667 & 4.026 & 4.591 & Q10 & \textbf{\underline{3.576}} & \textbf{\underline{3.894}} & \multicolumn{1}{c|}{\textbf{\underline{4.412}}} & Q10 & 3.221 & 3.465 & 3.894   & Q10 & \textbf{\underline{3.072}} & \textbf{\underline{3.368}} & \textbf{\underline{3.776}}     \\

 &   & Q50 & 7.808 & 8.499 & 9.862 & Q50 & \textbf{\underline{7.591}} & \textbf{\underline{8.228}} & \multicolumn{1}{c|}{\textbf{\underline{9.403}}} & Q50 & 6.788 & 7.207 & 7.964   & Q50 & \textbf{\underline{6.467}} & \textbf{\underline{6.932}} & \textbf{\underline{7.615}}     \\
  &   & Q90 & 3.695 & 4.086 & 4.964 & Q90 & \textbf{\underline{3.571}} & \textbf{\underline{3.959}} & \multicolumn{1}{c|}{\textbf{\underline{4.706}}} & Q90 & 3.190 & 3.452 & 3.907   & Q90 & \textbf{\underline{3.021}} & \textbf{\underline{3.282}} & \textbf{\underline{3.661}}     \\ \hline

\multirow{6}{*}{PEMSBAY} & \multirow{3}{*}{Point} & RMSE & 2.862 & 3.830 & 4.694 & RMSE & \textbf{\underline{2.800}} & \textbf{\underline{3.749}} & \multicolumn{1}{c|}{\textbf{\underline{4.549}}}  & RMSE & 2.788 & 3.652 & 4.318   & RMSE & \textbf{\underline{2.736}} & \textbf{\underline{3.595}} & \textbf{\underline{4.237}}        \\
 &   & MAE & 1.399 & 1.747 & 2.118 & MAE & \textbf{\underline{1.359}} & \textbf{\underline{1.703}} & \multicolumn{1}{c|}{\textbf{\underline{2.053}}} & MAE & 1.324 & 1.624 & 1.906   & MAE & \textbf{\underline{1.293}} & \textbf{\underline{1.588}} & \textbf{\underline{1.861}}    \\
 &   & MAPE & 3.013 & 3.978 & 5.034 & MAPE & \textbf{\underline{2.922}} & \textbf{\underline{3.899}} & \multicolumn{1}{c|}{\textbf{\underline{4.947}}} & MAPE & 2.839 & 3.691 & 4.510   & MAPE & \textbf{\underline{2.756}} & \textbf{\underline{3.595}} & \textbf{\underline{4.382}}        \\
& \multirow{3}{*}{Quantile} & Q10 & 0.380 & 0.487 & 0.618 & Q10 & \textbf{\underline{0.369}} & \textbf{\underline{0.475}} & \multicolumn{1}{c|}{\textbf{\underline{0.600}}} & Q10 & 0.930 & 1.113 & 1.367   & Q10 & \textbf{\underline{0.878}} & \textbf{\underline{1.045}} & \textbf{\underline{1.257}}     \\
 &   & Q50 & 0.700 & 0.873 & 1.059 & Q50 & \textbf{\underline{0.679}} & \textbf{\underline{0.851}} & \multicolumn{1}{c|}{\textbf{\underline{1.027}}} & Q50 & 0.662 & 0.812 & 0.953   & Q50 & \textbf{\underline{0.647}} & \textbf{\underline{0.794}} & \textbf{\underline{0.931}}     \\
 &   & Q90 & 0.301 & 0.364 & 0.431 & Q90 & \textbf{\underline{0.291}} & \textbf{\underline{0.354}} & \multicolumn{1}{c|}{\textbf{\underline{0.418}}} & Q90 & 0.282 & 0.338 & 0.391   & Q90 & \textbf{\underline{0.276}} & \textbf{\underline{0.333}} & \textbf{\underline{0.386}}     \\ \hline
\label{extra_experiment}
\end{tabular}
}
\end{table*}